\documentclass[10pt,onecolumn]{article}
\usepackage{times}
\usepackage[dvips]{graphicx}
\usepackage[hang,scriptsize]{subfigure}
\usepackage{latexsym}
\usepackage{amsmath} 
\usepackage{amssymb} 
\usepackage{graphics} 
\usepackage{listings}    
\usepackage{stmaryrd}    

\makeatletter
\def\@maketitle{%
  \vbox to 6.5cm{%
    \hsize\textwidth
    \linewidth\hsize
    \vspace{1.5cm}
    \centering
    {\bfseries\LARGE \@title \par}
    \vspace{24pt}
    {\fontsize{11pt}{13pt}\selectfont \begin{tabular}[t]{c}\@author \end{tabular}\par}
    \vfill} 
}
\renewcommand\section{\@startsection{section}{1}{\z@}%
                       {-12\p@ \@plus -4\p@ \@minus -4\p@}%
                       {6\p@ \@plus 4\p@ \@minus 4\p@}%
                       {\normalfont\large\bfseries
                        \rightskip=\z@ \@plus 8em\pretolerance=10000 }}
\renewcommand\subsection{\@startsection{subsection}{2}{\z@}%
                       {-12\p@ \@plus -4\p@ \@minus -4\p@}%
                       {6\p@ \@plus 4\p@ \@minus 4\p@}%
                       {\normalfont\fontsize{11pt}{13pt}\selectfont\bfseries
                        \rightskip=\z@ \@plus 8em\pretolerance=10000 }}
\renewcommand\subsubsection{\@startsection{subsubsection}{3}{\z@}%
                       {-12\p@ \@plus -4\p@ \@minus -4\p@}%
                       {6\p@ \@plus 4\p@ \@minus 4\p@}%
                       {\normalfont\normalsize\itshape}}
\renewcommand\paragraph{\@startsection{paragraph}{4}{\z@}%
                       {-12\p@ \@plus -4\p@ \@minus -4\p@}%
                       {-0.5em \@plus -0.22em \@minus -0.1em}%
                       {\normalfont\normalsize\itshape}}
\makeatother

\renewenvironment{abstract}%
  {\small
    \list{}{\labelwidth0pt
      \leftmargin0pt \rightmargin\leftmargin
      \listparindent\parindent \itemindent0pt
      \parsep0pt
      }%
    \item[\hskip\labelsep\bfseries\abstractname\enspace --] \itshape}{\endlist}

\newcommand{\keywordsname}{Keywords}
\newenvironment{keywords}%
  {\small
    \list{}{\labelwidth0pt
      \leftmargin0pt \rightmargin\leftmargin
      \listparindent\parindent \itemindent0pt
      \parsep0pt
      }%

    \item[\hskip\labelsep\bfseries\keywordsname:]}{\endlist}

\begin{document}

 \title{Proportional Conflict Redistribution Rules\\
 for Information Fusion}

\author{\begin{tabular}{c@{\extracolsep{4em}}c@{\extracolsep{1em}}c}
{\bf Florentin Smarandache} & {\bf Jean Dezert}\\
Dept. of Mathematics &  ONERA/DTIM/IED \\
Univ. of New Mexico  & 29 Av. de la  Division Leclerc \\
Gallup, NM 8730 & 92320 Ch\^{a}tillon \\
U.S.A.  & France \\
{\tt smarand@unm.edu} & {\tt Jean.Dezert@onera.fr}
\end{tabular}}

\date{}
\maketitle
\pagestyle{plain}

\begin{abstract}
In this paper we propose five versions of a Proportional Conflict Redistribution
rule (PCR) for information fusion together with several examples.
%
From PCR1 to PCR2, PCR3, PCR4, PCR5 one increases the complexity of the rules and also the exactitude of the redistribution of conflicting masses. PCR1 restricted from the hyper-power set to the power set and without degenerate cases gives the same result as the Weighted Average Operator (WAO) proposed recently by J{\o}sang, Daniel and Vannoorenberghe but does not satisfy the neutrality property of vacuous belief assignment. That's why improved PCR rules are proposed in this paper. PCR4 is an improvement of minC and Dempster's rules. The PCR rules redistribute the conflicting mass, after the conjunctive rule has been applied, proportionally with some functions depending on the masses assigned to their corresponding columns in the mass matrix. There are infinitely many ways these functions (weighting factors) can be chosen depending on the complexity one wants to deal with in specific applications and fusion systems. Any fusion combination rule is at some degree ad-hoc.
\end{abstract}


\section{Introduction}
This paper presents a new set of alternative combination rules based on different proportional conflict redistributions (PCR) which can be applied in the framework of the two principal theories dealing the combination of belief functions. We remind briefly the basic ideas of these two theories:
\begin{itemize}
\item The first and the oldest one is the Dempster-Shafer Theory (DST) developed by Shafer in 1976 in \cite{Shafer_1976}. In DST framework, Glenn Shafer starts with a so-called {\it{frame of discernment}} $\Theta=\{\theta_1,\ldots,\theta_n\}$ consisting in a finite set of exclusive and exhaustive hypotheses. This is the Shafer's model. Then, a basic belief assignment (bba) $m(.)$ is defined as the mapping $m: 2^\Theta\rightarrow [0, 1]$ with:
\begin{equation}
m(\emptyset)=0\qquad\text{and}\qquad
\displaystyle \sum_{X\in 2^\Theta} m(X)=1
\label{eq:bba}
\end{equation}
\noindent
The combination of belief assignments provided by several sources of evidence is done with the Dempster's rule of combination.
\item
The second and the most recent theory is the Dezert-Smarandache Theory (DSmT) developed by the authors since 2001 \cite{DSmTBook_2004a}. In the DSmT framework, one starts with a frame $\Theta=\{\theta_1,\ldots,\theta_n\}$ consisting only in a finite set of exhaustive\footnote{The exhaustivity assumption is not restrictive since one always can close any non-exhaustive set by introducing a closure element, say $\theta_0$, representing all missing unknown hypotheses.} hypotheses. This is the so-called {\it{free DSm model}}. The exclusivity assumption between elements (i.e. requirement for a refinement) of $\Theta$ is not necessary within DSmT. However, in DSmT any integrity constraints between elements of $\Theta$ can also be introduced, if necessary, depending on the fusion problem under consideration. A free DSm model including some integrity constraints is called a {\it{hybrid DSm model}}. DSmT can deal also with the Shafer's model as well which appears actually only as a specific hybrid DSm model. The DSmT framework is much larger that the DST one since it offers the possibility to deal with any model and any intrinsic nature of elements of $\Theta$ including continuous/vague concepts having subjective/relative interpretation which cannot be refined precisely into finer exclusive subsets. In DSmT, a generalized basic belief assignment (gbba) $m(.)$ is defined as the mapping $m: D^\Theta\rightarrow [0, 1]$ with \begin{equation}
m(\emptyset)=0\qquad\text{and}\qquad
\displaystyle \sum_{X\in D^\Theta} m(X)=1
\label{eq:gbba}
\end{equation}
\noindent
$D^\Theta$ represents the hyper-power set of $\Theta$ (i.e. Dedekind's lattice). Since the power set $2^\Theta$ is closed under $\cup$ operator, while the hyper-power set $D^\Theta$ is closed under both $\cup$ and $\cap$ operators, $\mid D^\Theta\mid > \mid 2^\Theta\mid$. A detailed presentation of DSmT with many examples and comparisons between rules of combination can be found in \cite{DSmTBook_2004a}.
\end{itemize}

Among all possible bbas or gbbas, the belief  vacuous belief assignment (VBA), denoted $m_v(.)$ and defined by $m_v(\Theta)=1$ which  characterizes a full ignorant source, plays a particular and important role for the construction of a satisfying combination rule. Indeed, the major properties that a good rule of combination must satisfy, upon to authors' opinion, are :
\begin{enumerate}
\item the coherence of the combination result in all possible cases (i.e. for any number of sources, any values of bbas or gbbas and for any types of frames and models which can change or stay invariant over time). 
\item the commutativity of the rule of combination
\item the neutral impact of the VBA into the fusion.
\end{enumerate}

The requirement for conditions 1 and 2 is legitimate since we are obviously looking for best performances (we don't want a rule yielding to counter-intuitive or wrong solutions) and we don't want that the result depends on the arbitrary order the sources are combined.
The neutral impact of VBA to be satisfied by a fusion rule (condition 3), denoted by the generic $\oplus$ operator is very important too. 
This condition states that the combination of a full ignorant source with a set of $s\geq 1$ non-totally ignorant sources doesn't change the result of the combination of the $s$ sources because the full ignorant source doesn't bring any new specific evidence on any problems under consideration. This condition is thus perfectly reasonable and legitimate. The condition 3 is mathematically represented as follows: for all possible $s\geq 1$ non-totally ignorant sources and for any $X\in 2^\Theta$ (or for any $X\in D^\Theta$ when working in the DSmT framework), the fusion operator $\oplus$ must satisfy
\begin{equation}
[m_1\oplus\ldots\oplus m_s\oplus m_v](X)=[m_1\oplus\ldots\oplus m_s](X)
\label{eq:VBAimpact}
\end{equation}

The associativity property, while very attractive and generally useful for sequential implementation is not actually a crucial property that a combination rule must satisfy if one looks for the best coherence of the result. The search for an optimal solution requires to process all bbas or gbbas altogether. Naturally, if several different rules of combination satisfy conditions 1-3 and provide similar performances, the simplest rule endowing associativity will be preferentially chosen (from engineering point of view). Up to now and unfortunately, no combination rule available in literature satisfy incontrovertibly the three first primordial conditions. Only three fusion rules based on the conjunctive operator are known associative: the Dempster's rule in DST, the SmetsÕ rule (conjunctive consensus based on the open-world assumption), and the DSm classic rule on free DSm model. The disjunctive rule is associative  and satisfy properties 1 and 2 only. All alternative rules developed in literature until now don't endow properties 1-3 and the associativity property. Although, some rules such as Yager's, Dubois \& Prade's, DSm hybrid, WAO, minC, PCR rules, which are not associative become quasi-associative if one stores the result of the conjunctive rule at each time when a new bba arises in the combination process. Instead of combining it with the previous result of the rule, we combine the new bba with the stored conjunctive rule's result. For unification of notations, we denote by $G$ either $2^\Theta$ or $D^\Theta$ depending on the theoretical framework chosen.\\

This paper extends a previous paper on Proportional Conflict Redistribution Rule no 1 (PCR1) detailed in \cite{Smarandache_Dezert_2004c} in order to overcome its inherent limitation (i.e. the neutral impact of VBA - condition 3 -  is not fulfilled by PCR1). In the DSm hybrid rule of combination, the transfer of partial conflicts (taking into account all integrity constraints of the model) is done directly onto the most specific sets including the partial conflicts but without proportional redistribution. In this paper, we propose to improve this rule by introducing a more effective proportional conflict redistribution to get a more efficient and precise rule of combination PCR5.\\

The main steps in applying all the PCR rules of combination (i.e. fusion) are as follows: 
\begin{itemize}
\item Step 1: use the conjunctive rule, 
\item Step 2: compute the conflicting masses (partial and/or total), 
\item Step 3: redistribute the conflicting masses to non-empty sets.
\end{itemize}

The way the redistribution is done makes the distinction between all existing rules available in literature in the DST and DSmT frameworks (to the knowledge of the authors) and the PCR
rules, and also the distinction among the different PCR versions themselves.
One also studies the impact of the vacuous belief assignment (VBA) on PCR rules and one makes a short discussion on the degree of the fusion rules' ad-hoc-ity. \\

Before presenting the PCR rules, and after a brief reminder on the notion of total and partial conflicts, we browse the main rules of combination proposed in the literature in the frameworks of DST and DSmT in the next section. Then we present the general Weighted Operator (WO), the Weighted Average Operator (WAO) and the minC operator. MinC is historically the first sophisticated rule using the idea of proportional conflict redistribution. The last part of this paper is devoted to the development of a new family of  PCR rules. Several examples and comparisons with other rules are also provided.

\section{The principal rules of combination}

In the sequel, we assume non degenerate void\footnote{The degenerate void problem considers $\Theta=\{\emptyset\}$ which is actually a meaningless fusion problem in static fusion applications since the frame contains no hypothese on which we can work with. In dynamic fusion application, a non degenerate void problem can sometimes turn into a degenerate void problem at a given time depending of the evolution of integrity constraints and thus the dynamic fusion problem can vanish with time. To overcome such possibility (if required by the fusion system designer), it is more cautious to always introduce at least one closure - possibly unknown - element $\theta_0\neq\emptyset$ in $\Theta$.} problems and thus we always consider the frame $\Theta$ as a truly non empty finite set (i.e. $\Theta\neq\{\emptyset\}$), unless specified expressly.

\subsection{Notion of  total and partial conflicting masses}

The {\it{total conflicting mass}} drawn from two sources, denoted $k_{12}$, is defined as follows:

\begin{equation}
k_{12}=\sum_{\substack{X_1,X_2\in G \\ X_1\cap X_2=\emptyset}}m_{1}(X_1)m_{2}(X_2)
\label{eq:TotalCM}
\end{equation}

The total conflicting mass is nothing but the sum of {\it{partial conflicting masses}}, i.e.

\begin{equation}
k_{12}=\sum_{\substack{X_1,X_2\in G \\ X_1\cap X_2=\emptyset}}m(X_1\cap X_2)
\label{eq:partialCM}
\end{equation}

Here, $m(X_1\cap X_2)$, where $X_1\cap X_2=\emptyset$, represents a partial conflict, i.e. the conflict between the sets $X_1$ and $X_2$.
Formulas \eqref{eq:TotalCM} and \eqref{eq:partialCM} can be directly generalized for $s \geq 2$ sources as follows:
\begin{equation}
k_{12\ldots s}=\sum_{\substack{X_1,\ldots,X_s\in G \\ X_1\cap \ldots\cap X_s=\emptyset}}\prod_{i=1}^{s}m_{i}(X_i)
\label{eq:TotalCMs}
\end{equation}

\begin{equation}
k_{12\ldots s}=\sum_{\substack{X_1,\ldots,X_s\in G \\  X_1\cap \ldots\cap X_s=\emptyset}}m(X_1\cap X_2\cap\ldots\cap X_s)
\label{eq:partialCMs}
\end{equation}

\subsection{The conjunctive rule}

\subsubsection{Definition}

For $n \geq 2$,  let's $\Theta= \{\theta_1, \theta_2,\ldots, \theta_n\}$ be the frame of the fusion problem under consideration. In the case when these $n$ elementary hypotheses $\theta_1, \theta_2,\ldots,\theta_n$ are known to be truly {\it{exhaustive and exclusive}} (i.e. the Shafer's model holds), one can use the DST \cite{Shafer_1976} framework with the Dempster's rule, the Yager's rule \cite{Yager_1985,Yager_1987}, the TBM \cite{Smets_1994,Smets_2000} approach, the Dubois-Prade approach \cite{Dubois_1986b,Dubois_1988,Dubois_1992} or the DSmT framework as well using the general DSm hybrid rule of combination \cite{DSmTBook_2004a} adapted to deal with any DSm model (including the Shafer's model). When the hypotheses (or some of them) are not exclusive and have potentially vague boundaries, the DSmT \cite{DSmTBook_2004a} is adopted. If hypotheses are known to be {\it{non-exhaustive}}, one can either uses either the Smets' open-world approach \cite{Smets_1994,Smets_2000} or apply the hedging closure procedure \cite{Yager_1983} and work back with DST or DSmT.\\

The conjunctive rule (known also as conjunctive consensus) for $s\geq2$ sources can be applied both in DST and in DSmT frameworks.
 In the DST framework, it is defined $\forall X \in 2^\Theta$ by
\begin{equation}
m_\cap(X)=\sum_{\substack{X_1,\ldots,X_s\in 2^\Theta \\ X_1\cap \ldots\cap X_s=X}}\prod_{i=1}^{s}m_{i}(X_i)
\label{eq:ConjPS}
\end{equation}
\noindent
$m_\cap(.)$ is not a proper belief assignment satisfying the Shafer's definition \eqref{eq:bba}, since in most of cases the sources do not totally agree (there exists partial and/or total conflicts between sources of evidence), so that $m_\cap(\emptyset) >0$. In Smets' open-world approach and TBM, one allows $m_\cap(\emptyset) \geq 0$ and the empty set is then interpreted not uniquely as the classical empty set (i.e. the set having no element) but also as the set containing all missing hypotheses of the original frame $\Theta$ to which the all conflicting mass is committed.\\

In the DSmT framework, the formula is similar, but instead of the power set $2^\Theta$, one uses the hyper-power set $D^\Theta$ and the generalized basic belief assignments, i.e. $\forall X \in D^\Theta$
\begin{equation}
m_\cap(X)=\sum_{\substack{X_1,\ldots,X_s\in D^\Theta \\ X_1\cap \ldots\cap X_s=X}}\prod_{i=1}^{s}m_{i}(X_i)
\label{eq:ConjHPS}
\end{equation}
$m_\cap(.)$ remains, in the DSmT framework based on the free DSm model, a proper generalized belief assignment as defined in \eqref{eq:gbba}. Formula \eqref{eq:ConjHPS} allowing the use of intersection of sets (for the non-exclusive hypotheses) is called the {\it{DSm classic rule}}.

\subsubsection{Example}

Let's consider $\Theta=\{\theta_1,\theta_2\}$ and two sources with belief assignments
$$m_1(\theta_1)=0.1\quad m_1(\theta_2)=0.2\quad m_1(\theta_1\cup \theta_2)=0.7$$
$$m_2(\theta_1)=0.4\quad m_2(\theta_2)=0.3\quad m_2(\theta_1\cup \theta_2)=0.3$$

\noindent
In the DST framework based on the Shafer's model, one gets
$$m_{\cap}(\emptyset)=0.11\quad m_{\cap}(\theta_1)=0.35$$
$$m_{\cap}(\theta_2)=0.33\quad m_{\cap}(\theta_1\cup \theta_2)=0.21$$
In the DSmT framework based on the free DSm model, one gets
$$m_{\cap}(\emptyset)=0\quad m_{\cap}(\theta_1\cap\theta_2)=0.11$$
$$m_{\cap}(\theta_1)=0.35\quad m_{\cap}(\theta_2)=0.33\quad m_{\cap}(\theta_1\cup \theta_2)=0.21$$

We can easily verify that the condition 3 (neutral impact of VBA) is satisfied with the conjunctive operator in both cases and that
the commutativity and associativity are also preserved. The main drawback of this operator is that it doesn't generate a proper belief assignment in both DST and DSmT frameworks when integrity constraints are introduced in the model as in dynamic fusion problems where the frame and/or the model itself can change with time.

\subsection{The disjunctive rule}

The disjunctive rule of combination \cite{Dubois_1986b,Dubois_1988,Smets_1993a} is a  commutative and associative rule proposed by Dubois \& Prade in 1986 and denoted here by the index $\cup$. $m_{\cup}(.)$ is defined $\forall X\in 2^\Theta$ by $m_{\cup}(\emptyset)=0$ and $ \forall (X\neq\emptyset) \in 2^\Theta$ by
$$m_{\cup}(X)=\displaystyle\sum_{\substack{X_1,X_2\in 2^\Theta\\ X_1\cup X_2=X}} m_{1}(X_1) m_{2}(X_2)$$

The core of the belief function (i.e. the set of focal elements having a positive mass) given by  $m_{\cup}$ equals the union of the cores of $m_1$ and $m_2$. 
This rule reflects the disjunctive consensus and is usually preferred when one knows that one of the sources (some of the sources in the case of $s$ sources) could be mistaken but without knowing which one. The disjunctive rule can also be defined similarly in DSmT framework by replacing $2^\Theta$ by $D^\Theta$ in the previous definition.

\subsection{The Dempster's rule }

The Dempster's rule of combination is the most widely used rule of combination so far in many expert systems based on belief functions since historically it was proposed in the seminal book of Shafer in \cite{Shafer_1976}. This rule, although presenting interesting advantages (mainly the commutativity, associativity and the neutral impact of VBA) fails however to provide coherent results due to the normalization procedure it involves. Some proponents of the Dempster's rule claim that this rule  provides correct and coherent result, but actually under strictly satisfied probabilistic conditions, which are rarely satisfied in common real applications. Discussions on the justification of the Dempster's rule and its well-known limitations can be found by example in \cite{Zadeh_1979,Zadeh_1984,Zadeh_1985,Voorbraak_1991,DSmTBook_2004a}. 
Let's a frame of discernment $\Theta$ based on the Shafer's model and two independent and equi-reliable belief assignments $m_1(.)$ and $m_2(.)$. The Dempster's rule of combination of $m_1(.)$ and $m_2(.)$ is obtained as follows: $m_{DS}(\emptyset)=0$ and $\forall (X\neq\emptyset) \in 2^\Theta$ by
\begin{equation}
m_{DS}(X) = \frac{\displaystyle\sum_{\substack{X_1,X_2\in 2^\Theta \\ X_1\cap X_2=X}}m_{1}(X_1)m_{2}(X_2)}
{\displaystyle 1-\sum_{\substack{X_1,X_2\in 2^\Theta\\ X_1\cap X_2=\emptyset}} m_{1}(X_1) m_{2}(X_2)}=\frac{1}{1-k_{12}}\cdot \displaystyle\sum_{\substack{X_1,X_2\in 2^\Theta \\ X_1\cap X_2=X}}m_{1}(X_1)m_{2}(X_2)
\label{eq:DSRule}
 \end{equation}
\noindent where the  {\it{degree of conflict}}  $k_{12}$ is defined by
$k_{12}\triangleq \displaystyle
\sum_{\substack{X_1,X_2\in 2^\Theta\\ X_1\cap X_2=\emptyset}} m_{1}(X_1) m_{2}(X_2)$.\\

\noindent
$m_{DS}(.)$ is a proper basic belief assignment if and only if the denominator in equation \eqref{eq:DSRule} is non-zero, i.e. the  {\it{degree of conflict}}  $k_{12}$ is less than one.

\subsection{The Smets' rule}

The Smets' rule of combination \cite{Smets_1994,Smets_2000} is nothing but the non-normalized version of the conjunctive consensus (equivalent to the  non-normalized version of Dempster's rule).  It is commutative and associative and allows positive mass on the null/empty set $\emptyset$ (i.e. open-world assumption). Smets' rule of combination of two independent (equally reliable) sources of evidence (denoted here by index $S$) is given by:

$$m_S(\emptyset)\equiv k_{12}=\displaystyle\sum_{\substack{X_1,X_2\in 2^\Theta\\ X_1\cap X_2=\emptyset}} m_{1}(X_1) m_{2}(X_2)$$

\noindent
and $ \forall (X\neq\emptyset) \in 2^\Theta$, by

$$m_S(X)=\displaystyle\sum_{\substack{X_1,X_2\in 2^\Theta \\ X_1\cap X_2=X}}m_{1}(X_1)m_{2}(X_2)$$

\subsection{The Yager's rule }

The Yager's rule of combination \cite{Yager_1983, Yager_1985,Yager_1987} admits that in case of conflict the result is not reliable, so that $k_{12}$ plays the role of an absolute discounting term added to the weight of ignorance. This commutative but not associative rule, denoted here by index $Y$ is given\footnote{$\Theta$ represents here the full ignorance $\theta_1\cup\theta_2\cup\ldots\cup\theta_n$ on the frame of discernment according the notation used in \cite{Shafer_1976}.} by $m_Y(\emptyset)=0$ and $ \forall X\in 2^\Theta, X\neq\emptyset,\! X\neq\Theta$ by
$$m_Y(X)=\displaystyle\sum_{\substack{X_1,X_2\in 2^\Theta\\ X_1\cap X_2=X}} m_1(X_1)m_2(X_2)$$
\noindent 
and when $X=\Theta$ by
$$m_Y(\Theta)=m_1(\Theta)m_2(\Theta) + \displaystyle\sum_{\substack{X_1,X_2\in 2^\Theta\\ X_1\cap X_2=\emptyset}} m_1(X_1)m_2(X_2)$$

\subsection{The Dubois \& Prade's rule}

The Dubois \& Prade's rule of combination \cite{Dubois_1988} admits that the two sources are reliable when they are not in conflict, but one of them is right when a conflict occurs. Then if one observes a value in set $X_1$ while the other observes this value in a set $X_2$, the truth lies in $X_1\cap X_2$ as long $X_1\cap X_2\neq \emptyset$. If $X_1\cap X_2=\emptyset$, then the truth lies in $X_1\cup X_2$ \cite{Dubois_1988}. According to this principle, the commutative (but not associative) Dubois \& Prade hybrid rule of combination, denoted here by index $DP$, which is a reasonable trade-off between precision and reliability, is defined
by $m_{DP}(\emptyset)=0$ and $ \forall X\in 2^\Theta, X\neq\emptyset$ by
\begin{equation}
m_{DP}(X)=\displaystyle\sum_{\substack{X_1,X_2\in 2^\Theta\\ X_1\cap X_2=X \\ X_1\cap X_2\neq\emptyset}} m_1(X_1)m_2(X_2)
+ \displaystyle\sum_{\substack{X_1,X_2\in 2^\Theta\\ X_1\cup X_2=X \\X_1\cap X_2=\emptyset}} m_1(X_1)m_2(X_2)
\end{equation}

\subsection{The hybrid DSm rule}

The hybrid DSm rule of combination is the first general rule of combination developed in the DSmT framework \cite{DSmTBook_2004a} which can work on any DSm models (including the Shafer's model) and for any level of conflicting information. The hybrid DSm rule can deal with the potential dynamicity of the frame and its model as well.
The DSmT deals properly with the granularity of information and intrinsic vague/fuzzy nature of elements of the frame $\Theta$ to manipulate. The basic idea of DSmT is to define belief assignments on hyper-power set $D^\Theta$ (i.e. free Dedekind's lattice) and to integrate all integrity constraints (exclusivity and/or non-existential constraints) of the model, say $\mathcal{M}(\Theta)$, fitting with the problem into the rule of combination. 
Mathematically, the hybrid DSm rule of combination of $s\geq 2$ independent sources of evidence is defined as follows (see chap. 4 in \cite{DSmTBook_2004a}) for all $X\in D^\Theta$,
\begin{equation}
m_{\mathcal{M}(\Theta)}(X)\triangleq 
\phi(X)\Bigl[ S_1(X) + S_2(X) + S_3(X)\Bigr]
 \label{eq:DSmHkBis1}
\end{equation}
\noindent
where $\phi(X)$ is the {\it{characteristic non-emptiness function}} of a set $X$, i.e. $\phi(X)= 1$ if  $X\notin \boldsymbol{\emptyset}$ and $\phi(X)= 0$ otherwise, where $\boldsymbol{\emptyset}\triangleq\{\boldsymbol{\emptyset}_{\mathcal{M}},\emptyset\}$. $\boldsymbol{\emptyset}_{\mathcal{M}}$ is the set  of all elements of $D^\Theta$ which have been forced to be empty through the constraints of the model $\mathcal{M}$ and $\emptyset$ is the classical/universal empty set. $S_1(X)$, $S_2(X)$ and $S_3(X)$ are defined by 
\begin{equation}
S_1(X)\triangleq \sum_{\substack{X_1,X_2,\ldots,X_s\in D^\Theta\\ (X_1\cap X_2\cap\ldots\cap X_s)=X}} \prod_{i=1}^{s} m_i(X_i)
\end{equation}
\begin{equation}
S_2(X)\triangleq \sum_{\substack{X_1,X_2,\ldots,X_s\in\boldsymbol{\emptyset}\\ [\mathcal{U}=X]\vee [(\mathcal{U}\in\boldsymbol{\emptyset}) \wedge (X=I_t)]}} \prod_{i=1}^{s} m_i(X_i)
\end{equation}
\begin{equation}
S_3(A)\triangleq\sum_{\substack{X_1,X_2,\ldots,X_k\in D^\Theta \\ u(c(X_1\cap X_2\cap\ldots\cap X_k))=A \\ (X_1\cap X_2\cap \ldots\cap X_k)\in\boldsymbol{\emptyset}}}  \prod_{i=1}^{k} m_i(X_i)
\end{equation}
with $\mathcal{U}\triangleq u(X_1)\cup u(X_2)\cup \ldots \cup u(X_k)$ where $u(X)$ is the union of all $\theta_i$ that compose $X$, $I_t \triangleq \theta_1\cup \theta_2\cup\ldots\cup \theta_n$ is the total ignorance, and $c(X)$ is the canonical form\footnote{The canonical form is introduced here in order to improve the original formula given in \cite{DSmTBook_2004a} for preserving the neutral impact of the vacuous belief mass $m(\Theta)=1$ within complex hybrid models. The canonical form is the conjunctive normal form, also known as conjunction of disjunctions in Boolean algebra, which is unique.} of $X$, i.e. its simplest form (for example if $X=(A\cap B)\cap (A\cup B\cup C)$, 
$c(X)=A\cap B$. $S_1(A)$ corresponds to the classic DSm rule for $k$ independent sources based on the free DSm model $\mathcal{M}^f(\Theta)$; $S_2(A)$ represents the mass of all relatively and absolutely empty sets which is transferred to the total or relative ignorances associated with non existential constraints (if any, like in some dynamic problems); $S_3(A)$ transfers the sum of relatively empty sets directly onto the canonical disjunctive form of non-empty sets. The hybrid DSm rule generalizes the classic DSm rule of combination and is not equivalent to Dempster's rule. It works for any DSm models (the free DSm model, Shafer's model or any other hybrid models) when manipulating {\it{precise}} generalized (or eventually classical) basic belief functions. Extension of this hybrid DSm rule for the fusion of imprecise belief can be found in \cite{DSmTBook_2004a}.\\

In the case of a dynamic fusion problem, when all elements become empty because one gets new evidence on integrity constraints (which corresponds to a specific hybrid model $\mathcal{M}$), then the conflicting mass is transferred to the total ignorance, which also turns to be empty, therefore 
the empty set gets now mass equals one which shows that the problem has no solution at all (actually the problem is a degenerate void problem since all elements became empty at a given time). If we prefer to adopt an optimistic vision, we can consider that one (or more missing hypotheses), say $\theta_0$, has entered in the frame but we did pay attention to it in the dynamicity and thus, one must expressly consider $m(\theta_0)=1$ instead of $m(\emptyset)=1$. For example, Let's consider the frame $\Theta=\{A,B\}$ with the 2 following bbas $m_1(A)=0.5$, $m_1(B)=0.3$, $m_1(A\cup B)=0.2$ and $m_2(A)=0.4$, $m_2(B)=0.5$, $m_2(A\cup B)=0.1$, but one finds out with new evidence that $A$ and $B$ are truly empty, then 
$A\cup B\equiv\Theta\overset{\mathcal{M}}{\equiv}\emptyset$. Then $m(\emptyset)=1$ which means that this is a totally impossible problem because this degenerate problem turns out to be void. The only escape is to include a third or more missing hypotheses $C$, $D$, etc into the frame to warranty its true closure.\\

The hybrid DSm rule of combination is not equivalent to Dempster's rule even working on the Shafer's model. DSmT is an extension of DST in the way that the hyper-power set is an extension of the power set; hyper-power set includes, besides, unions, also intersections of elements; and when all intersections are empty, the hyper-power set coincides with the power set.  Consequently, the DSm hybrid models include the Shafer's model. An extension of this rule for the combination of {\it{imprecise}} generalized (or eventually classical) basic belief functions is possible and is presented in \cite{DSmTBook_2004a}. The hybrid DSm rule can be seen as an improved version of Dubois \& Prade's rule which mix the conjunctive and disjunctive consensus applied in the DSmT framework to take into account the possibility for any dynamical integrity constraint in the model.

\section{The general weighted operator (WO)}

In the framework of Dempster-Shafer Theory (DST), an unified formula has been proposed recently by Lef\`evre, Colot and Vanoorenberghe in \cite{Lefevre_2002} to embed all the existing (and potentially forthcoming) combination rules involving conjunctive consensus in the same general mechanism of construction. It turns out that such unification formula had been already proposed by Inagaki \cite{Inagaki_1991} in 1991 as reported in \cite{Sentz_2002}. This formulation is known as {\it{the Weighted Operator}} (WO) in literature \cite{Josang_2003}. The WO for 2 sources is based on two steps.

\begin{itemize}
\item {\bf{Step 1}}: Computation of the total conflicting mass based on the conjunctive consensus
\begin{equation}
k_{12} \triangleq \displaystyle\sum_{\substack{X_1,X_2\in 2^\Theta\\ X_1\cap X_2=\emptyset}}m_1(X_1)m_2(X_2) \end{equation}
\item {\bf{Step 2}}: This second step consists in the reallocation (convex combination) of the conflicting masses on $(X\neq\emptyset)\subseteq \Theta$ with some given coefficients $w_m(X)\in[0,1]$ such that $\sum_{X\subseteq \Theta} w_m(X)=1$ according to
$$m(\emptyset)= w_m(\emptyset)\cdot k_{12}$$
\noindent
and $\forall (X\neq\emptyset)\in 2^\Theta$
\begin{equation}
m(X) = [\displaystyle\sum_{\substack{X_1,X_2\in 2^\Theta\\ X_1\cap X_2=X}} m_1(X_1)m_2(X_2)] + w_m(X)k_{12}
\label{eq:ILCV}
\end{equation}
\end{itemize}

The WO can be easily generalized for the combination of $s\geq 2$ independent and equi-reliable sources of information as well  by substituting $k_{12}$ in step 1 by 
$$k_{12\ldots s}\triangleq \displaystyle\sum_{\substack{X_1,\ldots,X_s\in 2^\Theta\\ X_1\cap\ldots\cap X_s=\emptyset}}\prod_{i=1}^{s}m_i(X_i)$$
\noindent
and for step 2 by deriving for all $(X\neq\emptyset)\in 2^\Theta$ the mass $m(X)$ by
\begin{equation*}
m(X) = [\displaystyle\sum_{\substack{X_1,\ldots,X_s\in 2^\Theta\\ X_1\cap\ldots\cap X_s=X}} \prod_{i=1}^{s}m_i(X_i)] + w_m(X)k_{12\ldots s}
\end{equation*}

The particular choice of coefficients $w_m(.)$ provides a particular rule of combination (Dempster's, Yager's, Smets', Dubois \& Prade's rules, by example, are particular cases of WO \cite{Lefevre_2002}).
Actually this nice and important general formulation shows there exists an infinite number of possible rules of combination. Some rules are more justified or criticized with respect to the other ones mainly on their ability to, or not to, preserve the commutativity, associativity of the combination, to maintain the neutral impact of VBA and to provide what we feel coherent/acceptable solutions in high conflicting situations. It can be easily shown in  \cite{Lefevre_2002}  that such general procedure provides all existing rules involving conjunctive consensus developed in the literature based on Shafer's model. 

\section{The weighted average operator (WAO)}

\subsection{Definition}

This operator has been recently proposed (only in the framework of the Dempster-Shafer theory) by J{\o}sang, Daniel and Vannoorenberghe in \cite{Josang_2003} only for static fusion case. It is a new particular case of WO where the weighting coefficients $w_m(A)$ are chosen as follows: $w_m(\emptyset)=0$ and $\forall X\in 2^\Theta\setminus\{\emptyset\}$,
\begin{equation}
w_m(X)=\frac{1}{s} \sum_{i=1}^{s} m_i(X)
\label{eq:wm_WAO}
\end{equation}
\noindent
where $s$ is the number of independent sources to combine. \\

From the general expression of WO and this particular choice of weighting coefficients $w_m(X)$, one gets, for the combination of $s\geq 2$ independent sources and  $\forall (X\neq\emptyset)\in 2^\Theta$
\begin{equation}
m_{WAO}(X) = [\displaystyle\sum_{\substack{X_1,\ldots,X_s\in 2^\Theta\\ X_1\cap\ldots\cap X_s=X}} \prod_{i=1}^{s} m_i(X_i) ] 
+ [\frac{1}{s} \sum_{i=1}^{s} m_i(X)]\cdot
[\displaystyle\sum_{\substack{X_1,\ldots,X_s\in 2^\Theta\\ X_1\cap \ldots\cap X_s=\emptyset}} \prod_{i=1}^{s} m_i(X_i) ]
\label{eq:WAO}
\end{equation}

\subsection{Example for WAO}
\label{SectionWAOEx}
 Let's consider the Shafer's model (exhaustivity and exclusivity of hypotheses) on $\Theta=\{A,B\}$ and the two following bbas 
$$m_1(A)=0.3 \quad m_1(B)=0.4\quad m_1(A\cup B)=0.3$$
$$m_2(A)=0.5\quad m_2(B)=0.1\quad m_2(A\cup B)=0.4$$
\noindent The conjunctive consensus yields\footnote{We use $m_{12}$ instead of $m_{\cap}$ to indicate explicitly that only 2 sources enter in the conjunctive operator. The notation $m_{WAO|12}$ denotes the result of the WAO combination for sources 1 and 2. When 
$s\geq 2$ sources are combined, we use similarly the notations $m_{12\ldots s}$ and $m_{WAO|12\ldots s}$.}
$$m_{12}(A)=0.42\quad m_{12}(B)=0.23\quad m_{12}(A\cup B)=0.12$$
\noindent
with the conflicting mass $k_{12}=0.23$. The weighting average coefficients are given by
$$w_m(A)=0.40\quad w_m(B)=0.25\quad w_m(A\cup B)=0.35$$
The result of the WAO is therefore given by
\begin{align*}
m_{WAO|12}(A)&=m_{12}(A) + w_m(A)\cdot k_{12}=0.42+ 0.40\cdot 0.23=0.5120\\
m_{WAO|12}(B)&=m_{12}(B) + w_m(B)\cdot k_{12}=0.23+ 0.25\cdot 0.23=0.2875\\
m_{WAO|12}(A\cup B)&=m_{12}(A\cup B) + w_m(A\cup B)\cdot k_{12}=0.12+ 0.35\cdot 0.23=0.2005
\end{align*}

\subsection{Limitations of WAO}

From the previous simple example, one can easily verify that the WAO doesn't preserve the neutral impact of VBA (condition expressed in  \eqref{eq:VBAimpact}). Indeed, if one combines the two first sources with a third (but totally ignorant) source represented by the vacuous belief assignment  (i.e. $m_3(.)=m_v(.)$), $m_3(A\cup B)=1$ altogether, one gets same values from conjunctive consensus and conflicting mass, i.e. $k_{123}=0.23$ and
$$m_{123}(A)=0.42\quad m_{123}(B)=0.23\quad m_{123}(A\cup B)=0.12$$
but the weighting average coefficients are now given by
$$w_m(A)=0.8/3\quad w_m(B)=0.5/3\quad w_m(A\cup B)=1.7/3$$
so that
\begin{equation*}
m_{WAO|123}(A)=0.42+ (0.8/3)\cdot 0.23\approx 0.481333
\end{equation*}
\begin{equation*}
m_{WAO|123}(B)=0.23+ (0.5/3)\cdot 0.23\approx 0.268333
\end{equation*}
\begin{equation*}
m_{WAO|123}(A\cup B)=0.12+ (1.7/3)\cdot 0.23\approx 0.250334
\end{equation*}
Consequently,  WAO doesn't preserve the neutral impact of VBA since one has found at least one example in which condition
\eqref{eq:VBAimpact} is not satisfied because
\begin{equation*}
m_{WAO|123}(A)\neq m_{WAO|12}(A)
\end{equation*}
\begin{equation*}
m_{WAO|123}(B)\neq m_{WAO|12}(B)
\end{equation*}
\begin{equation*}
m_{WAO|123}(A\cup B)\neq m_{WAO|12}(A\cup B)
\end{equation*}

Another limitation of WAO concerns its impossibility to deal with dynamical evolution of the frame (i.e. when some evidence arises after a while on the true vacuity of elements of power set). As example, let's consider three different suspects $A$, $ B$ and $C$ in a criminal investigation (i.e. $\Theta=\{A,B,C\}$) and the two following simple Bayesian witnesses reports
$$m_1(A)=0.3 \quad m_1(B)=0.4\quad m_1(C)=0.3$$
$$m_2(A)=0.5\quad m_2(B)=0.1\quad m_2(C)=0.4$$
\noindent The conjunctive consensus is 
$$m_{12}(A)=0.15\quad m_{12}(B)=0.04\quad m_{12}(C)=0.12$$
\noindent
with the conflicting mass $k_{12}=0.69$.
Now let's assume that a little bit later, one learns that $B=\emptyset$ because the second suspect brings a perfect alibi, then the initial consensus on $B$ (i.e. $m_{12}(B)=0.04$) must enter now in the new conflicting mass $k_{12}'=0.69+0.04=0.73$ since $B=\emptyset$. $k_{12}'$ is then re-distributed to $A$ and $C$ according to the WAO formula:
$$m_{WAO|12}(B) = 0$$
$$m_{WAO|12}(A) = 0.15 + (1/2)(0.3+0.5)(0.73) = 0.4420$$
$$m_{WAO|12}(C) = 0.12 + (1/2)(0.3+0.4)(0.73) = 0.3755$$
\noindent
From this WAO result, one sees clearly that the sum of the combined belief assignments $m_{WAO|12}(.)$ is $0.8175 < 1$.
Therefore, the WAO proposed in \cite{Lefevre_2002} doesn't manage properly the combination with VBA neither the  possible dynamicity of the fusion problematic. This limitation is not very surprising since the WAO was proposed actually only for the static fusion\footnote{Note that the static fusion aspect was not explicitly stated and emphasized in \cite{Lefevre_2002} but only implicitly assumed.} based on Shafer's model. The improvement of WAO for dynamic fusion is an open problem, but Milan Daniel in a private communication to the authors, proposed to use the following normalized coefficients for WAO in dynamic fusion:
\begin{equation}
w_m(X)=\frac{1}{s} \frac{\sum_{X}\sum_{i=1}^{s} m_i(X)}{\sum_{X\neq\emptyset}\sum_{i=1}^{s} m_i(X)} \sum_{i=1}^{s} m_i(X)
\label{eq:wm_WAODyn}
\end{equation}
\noindent

\section{The Daniel's minC rule}

\subsection{Principle of the minC rule}
MinC fusion rule is a recent interesting rule based on proportional redistribution of partial conflicts. Actually it was the first rule, to the knowledge of authors, that uses the idea for sophisticated proportional conflict redistribution. This rule was developed in the DST framework only. MinC rule is commutative and preserves the neutral impact of VBA but, as the majority of rules, MinC is not fully associative. MinC has been developed and proposed by Milan Daniel in \cite{Daniel98x,Daniel_2000,Daniel_2000b,Daniel_2003}. A detailed presentation of MinC can also be found in \cite{DSmTBook_2004a} (Chap. 10). \\

The basic idea of minC is to identify all different types of partial conflicts and then transfer them with some proportional redistribution. Two versions of proportional redistributions have been proposed by Milan Daniel:
\begin{itemize}
\item The minC (version a) ): the mass coming from a partial conflict (called contradiction by M. Daniel) involving several sets $X_1$,$X_2$,\ldots,$X_k$ is proportionalized among all unions $\bigcup_{i=1}^j$, of $j\leq k$ sets $X_i$ of $\{X_1,\ldots,X_k\}$ (after a proper reallocation of all equivalent propositions containing partial conflit onto elements of power set).
\item The minC (version b) ): the mass coming from a partial conflict involving several sets $X_1$,$X_2$,\ldots,$X_k$ is proportionalized among all non empty subsets of $X_1\cup,\ldots\cup X_k$.
\end{itemize}
The preservation of the neutral impact of the VBA by minC rule can been drawn from the following demonstration: Let's consider two basic belief assignments $m_1(.)$ and $m_2(.)$. The first stage of minC consists in deriving the conjunctive consensus $m_{12}(.)$ from $m_1(.)$ and $m_2(.)$ and then transfer the mass of conflicting propositions to its components and unions of its components proportionally to their masses $m_{12}(.)$. Since the vacuous belief assignment $m_v(.)$ is the neutral element of the conjunctive operator, one always has $m_{12v}(.)=m_{12}(.)$ and thus the result of the minC at the first stage and after the first stage not affected by the introduction of the vacuous belief assignment in the fusion process. That's why minC preserves the neutral impact of VBA.\\

Unfortunately no analytic expression for the minC rules (version a and b) has been provided so far by the author. As simply stated, 
minC transfers $m(A\cap B)$ when $A\cap B=\emptyset$ with specific proportionalization factors to $A$, $B$, and $A\cup B$; More generally, minC transfers the conflicting mass $m(X)$, when $X=\emptyset$, to all subsets of $u(X)$ (the disjunctive form of $X$), which is not the most exact issue. As it will be shown in the sequel of this paper, the PCR5 rule allows a more judicious proportional conflict redistribution. For a better understanding of the minC rule, here is a simple illustrative example drawn from  \cite{DSmTBook_2004a} (p. 237).

\subsection{Example for minC}

Let's consider the Shafer's model with $\Theta=\{\theta_1,\theta_2,\theta_3\}$ and the two following bbas to combine (here we denotes $\theta_1\cup\theta_2\cup\theta_3$ by $\Theta$ for notation convenience).
\begin{align*}
m_1(\theta_1)& =0.3 & m_2(\theta_1)&=0.1 \\
m_1(\theta_2)& =0.2 & m_2(\theta_2)&=0.1 \\
m_1(\theta_3)& =0.1 & m_2(\theta_3)&=0.2 \\
m_1(\theta_1\cup\theta_2)& =0.1 & m_2(\theta_1\cup\theta_2)&=0.0 \\
m_1(\theta_1\cup\theta_3)& =0.1 & m_2(\theta_1\cup\theta_3)&=0.1 \\
m_1(\theta_2\cup\theta_3)& =0.0 & m_2(\theta_2\cup\theta_3)&=0.2 \\
m_1(\Theta)& =0.2 & m_2(\Theta)&=0.3 
\end{align*}
The results of the three steps of the minC rules are given in Table \ref{Table1}. For notation convenience, the square symbol $\square$ represents  $(\theta_1\cap\theta_2)\cup(\theta_1\cap\theta_3)\cup(\theta_2\cap\theta_3)$. 

\begin{table}[h]
\begin{center}
\begin{tabular}{|c||c|c|c|c|}
\hline
      & $m_{12}$ & $m_{12}^\star$ & $m_{\text{\tiny minC}}^a$ & $m_{\text{\tiny minC}}^b$  \\ 
\hline 
$\theta_1$ & 0.19 & 0.20 & 0.2983 & 0.2999\\
$\theta_2$ & 0.15 & 0.17 & 0.2318 & 0.2402\\
$\theta_3$ & 0.14 & 0.16 & 0.2311 & 0.2327\\
$\theta_1\cup\theta_2$ & 0.03 & 0.03 & 0.0362 & 0.0383\\
$\theta_1\cup\theta_3$ & 0.06 & 0.06 & 0.0762 & 0.0792\\
$\theta_2\cup\theta_3$ & 0.04 & 0.04 & 0.0534 & 0.0515\\
$\theta_1\cup\theta_2\cup\theta_3$ & 0.06 & 0.06 & 0.0830 & 0.0692\\
\hline
$\theta_1\cap\theta_2$ & 0.05 & 0.05 &  & \\
$\theta_1\cap\theta_3$ & 0.07 & 0.07 &  & \\
$\theta_2\cap\theta_3$ & 0.05 & 0.05 &  & \\
$\theta_1\cap(\theta_2\cup\theta_3)$ & 0.06 & 0.06 &  & \\
$\theta_2\cap(\theta_1\cup\theta_3)$ & 0.03 & 0.03 &  & \\
$\theta_3\cap(\theta_1\cup\theta_2)$ & 0.02 & 0.02 &  & \\
$\theta_1\cup(\theta_2\cap\theta_3)$ & 0.01 &  &  & \\
$\theta_2\cup(\theta_1\cap\theta_3)$ & 0.02 &  &  & \\
$\theta_3\cup(\theta_1\cap\theta_2)$ & 0.02 &  &  & \\
$\theta_1\cap\theta_2\cap\theta_3$ & 0 &  &  & \\
$\square$ & 0 &  &  & \\
\hline\end{tabular}
\end{center}
\caption{minC result (versions a and b)}
\label{Table1}
\end{table}

\begin{itemize}
\item {\bf{Step 1 of minC}} : the conjunctive consensus\\

The first column of Table \ref{Table1} lists all the elements involved in the combination.
The second column gives the result of the first step of the minC rule which consists in applying the conjunctive consensus operator $m_{12}(.)$ defined on the hyper-power set $D^\Theta$ of the free-DSm model. 

\item {\bf{Step 2 of minC}} : the reallocation\\

The second step of minC consists in the reallocation of the masses of all partial conflicts which are equivalent to some non empty elements of the power set. This is what we call {\it{the equivalence-based reallocation principle}} (EBR principle).
The third column $m_{12}^\star$ of Table \ref{Table1} gives the basic belief assignment {\it{after reallocation of partial conflicts}} based on EBR principle before proportional conflict redistribution (i.e. the third and final step of minC). \\

Let's explain a bit what EBR is from this simple example. Because we are working with the Shafer's model all elements $\theta_1$, $\theta_2$ and $\theta_3$ of $\Theta$ are exclusive and therefore $\theta_1\cap\theta_2=\emptyset$, $\theta_1\cap\theta_3=\emptyset$, $\theta_3\cap\theta_3=\emptyset$ and $\theta_1\cap\theta_2\cap\theta_3=\emptyset$. Consequently, the propositions $\theta_1\cup(\theta_2\cap\theta_3)$, $\theta_2\cup(\theta_1\cap\theta_3)$, and $\theta_3\cup(\theta_1\cap\theta_2)$ corresponding to the 14th, 15th and 16th rows of the Table \ref{Table1} are respectively equivalent to $\theta_1$, $\theta_2$ and $\theta_3$ so that their committed masses can be directly reallocated (added) onto $m_{12}(\theta_1)$, $m_{12}(\theta_2)$ and $m_{12}(\theta_3)$. No other mass containing partial conflict can be directly reallocated onto the first seven elements of the table based on the EBR principle in this example. Thus finally, one gets $m_{12}^\star(.)=m_{12}(.)$ for all non-equivalent elements and for elements $\theta_1$, $\theta_2$ and $\theta_3$ for which a reallocation has been done
\begin{align*}
m_{12}^\star(\theta_1)&=m_{12}(\theta_1) + m_{12}(\theta_1\cup(\theta_2\cap\theta_3))=0.19+0.01=0.20\\
m_{12}^\star(\theta_2)& =m_{12}(\theta_2) + m_{12}(\theta_2\cup(\theta_1\cap\theta_3))=0.15+0.02=0.17\\
m_{12}^\star(\theta_3)&=m_{12}(\theta_3) + m_{12}(\theta_3\cup(\theta_1\cap\theta_2))=0.14+0.02=0.16
\end{align*}

\item {\bf{Step 3 of minC}} : proportional conflict redistribution\\

The fourth and fifth columns of the Table \ref{Table1} ($m_{\text{minC}}^a$ and $m_{\text{minC}}^b$) provide the minC results with the two versions of minC proposed by Milan Daniel and explicated below. The column 4 of the Table \ref{Table1} corresponds to the version a) of minC while the column 5 corresponds to the version b).
Let's explain now in details how the values of columns 4 and 5 have be obtained.\\

\noindent
{\it{Version a) of minC}}: The result for the minC (version a) corresponding to the fourth column of the Table \ref{Table1} is obtained from $m_{12}^\star(.)$ by the  proportional redistribution of the partial conflict onto the elements entering in the partial conflict and their union. By example, the mass $m_{12}^\star(\theta_1\cap(\theta_2\cup\theta_3))=0.06$ will be proportionalized from the mass of $\theta_1$, $\theta_2\cup\theta_3$ and $\theta_1\cup \theta_2\cup\theta_3$ only. The parts of the mass of $\theta_1\cap(\theta_2\cup\theta_3)$ added to $\theta_1$, $\theta_2\cup\theta_3$ and $\theta_1\cup \theta_2\cup\theta_3$ will be given by
\begin{align*}
k(\theta_1)&= m_{12}^\star(\theta_1\cap(\theta_2\cup\theta_3))\cdot \frac{m_{12}^\star(\theta_1)}{K}=0.06\cdot\frac{0.20}{0.30}=0.040\\
k(\theta_2\cup\theta_3) &= m_{12}^\star(\theta_1\cap(\theta_2\cup\theta_3))\cdot \frac{m_{12}^\star(\theta_2\cup\theta_3)}{K} =0.06\cdot\frac{0.04}{0.30}=0.008\\
k(\theta_1\cup \theta_2\cup\theta_3)&= m_{12}^\star(\theta_1\cap(\theta_2\cup\theta_3))\cdot \frac{m_{12}^\star(\Theta)}{K}=0.06\cdot\frac{0.06}{0.30}=0.012
\end{align*}
\noindent
where the normalization constant is $K=m_{12}^\star(\theta_1)+m_{12}^\star(\theta_2\cup\theta_3)+m_{12}^\star(\theta_1\cup\theta_2\cup\theta_3)=0.20+0.04+0.06=0.30$.

The proportional redistribution is done similarly for all other partial conflicting masses. We summarize in Tables 2-4 all the proportions (rounded at the fifth decimal) of conflicting masses to transfer onto elements of the power set.
The sum of each column of the Tables \ref{Table2}-\ref{Table4} is transferred onto the mass of the element of power set  it corresponds to get the final result of minC (version a)). By example, $m_{\text{\tiny minC}}^a(\theta_1)$ is obtained by
\begin{align*}
m_{\text{\tiny minC}}^a(\theta_1)& = m_{12}^\star(\theta_1)+ (0.025+0.03333+0.04)= 0.20+0.09833=0.29833
\end{align*}
\noindent
which corresponds to the first value (rounded at the 4th decimal) of the 4th column of Table \ref{Table1}.
All other values of the minC (version a) result of Table \ref{Table1} can be easily verified similarly.

\begin{table}[h]
\begin{center}
\begin{tabular}{|c||c|c|c|}
\hline
    & $\theta_1$ & $\theta_2$ & $\theta_3$ \\ 
\hline 
$\theta_1\cap\theta_2$ & 0.025 & 0.02125 &  \\
$\theta_1\cap\theta_3$ & 0.03333 &  & 0.02667 \\
$\theta_2\cap\theta_3$ &  & 0.02297 & 0.02162\\
$\theta_1\cap(\theta_2\cup\theta_3)$ & 0.04 &  &  \\
$\theta_2\cap(\theta_1\cup\theta_3)$ &  & 0.01758 & \\
$\theta_3\cap(\theta_1\cup\theta_2)$ &  &  & 0.0128 \\
\hline\end{tabular}
\caption{Version a) of minC Proportional conflict redistribution factors}
\label{Table2}
\end{center}
\end{table}
\begin{table}[h]
\begin{center}
\begin{tabular}{|c||c|c|}
\hline
    & $\theta_1\cup\theta_2$ & $\theta_1\cup\theta_3$ \\ 
\hline 
$\theta_1\cap\theta_2$ & 0.00375 &  \\
$\theta_1\cap\theta_3$ &  &  0.01 \\
$\theta_2\cap\theta_3$ &  &  \\
$\theta_1\cap(\theta_2\cup\theta_3)$ &  &  \\
$\theta_2\cap(\theta_1\cup\theta_3)$ &  &  0.00621 \\
$\theta_3\cap(\theta_1\cup\theta_2)$ & 0.0024 & \\
\hline\end{tabular}
\caption{Version a) of minC Proportional conflict redistribution factors (continued)}
\label{Table3}
\end{center}
\end{table}
\begin{table}[h]
\begin{center}
\begin{tabular}{|c||c|c|}
\hline
    &  $\theta_2\cup\theta_3$ & $\theta_1\cup\theta_2\cup\theta_3$\\ 
\hline 
$\theta_1\cap\theta_2$ &  & \\
$\theta_1\cap\theta_3$ &  & \\
$\theta_2\cap\theta_3$ & 0.00541 &\\
$\theta_1\cap(\theta_2\cup\theta_3)$ &   0.008 & 0.012\\
$\theta_2\cap(\theta_1\cup\theta_3)$ & & 0.00621\\
$\theta_3\cap(\theta_1\cup\theta_2)$ & & 0.0048 \\
\hline\end{tabular}
\caption{Version a) of minC Proportional conflict redistribution factors (continued)}
\label{Table4}
\end{center}
\end{table}

\noindent
{\it{Version b) of minC}}: In this second version of minC, the proportional redistribution of any partial conflict $X$ remaining after step 2 uses all subsets of $u(X)$ (i.e. the disjunctive form of $X$). As example, let's consider the partial conflict $X=\theta_1\cap(\theta_2\cup\theta_3)$ in the Table \ref{Table1} having the belief mass $m_{12}^\star(\theta_1\cap(\theta_2\cup\theta_3))=0.06$. Since $u(X)=\theta_1\cup\theta_2\cup\theta_3$, all elements of the power set $2^\Theta$ will enter in the proportional redistribution and we will get for this $X$
\begin{align*}
k(\theta_1)&=m_{12}^\star(\theta_1\cap(\theta_2\cup\theta_3))\cdot\frac{m_{12}^\star(\theta_1)}{K}
\approx 0.01666\\
k(\theta_2)&=m_{12}^\star(\theta_1\cap(\theta_2\cup\theta_3))\cdot\frac{m_{12}^\star(\theta_2)}{K}
\approx 0.01417\\
k(\theta_3)&=m_{12}^\star(\theta_1\cap(\theta_2\cup\theta_3))\cdot\frac{m_{12}^\star(\theta_3)}{K}
\approx 0.01333\\
k(\theta_1\cup\theta_2)&=m_{12}^\star(\theta_1\cap(\theta_2\cup\theta_3))\cdot\frac{m_{12}^\star(\theta_1\cup\theta_2)}{K}=0.06\cdot \frac{0.03}{0.72}=0.0025\\
k(\theta_1\cup\theta_3)&=m_{12}^\star(\theta_1\cap(\theta_2\cup\theta_3))\cdot\frac{m_{12}^\star(\theta_1\cup\theta_3)}{K}=0.06\cdot \frac{0.06}{0.72}=0.005\\
k(\theta_2\cup\theta_3)&=m_{12}^\star(\theta_1\cap(\theta_2\cup\theta_3))\cdot\frac{m_{12}^\star(\theta_2\cup\theta_3)}{K}=0.06\cdot \frac{0.04}{0.72}\approx 0.00333\\
k(\Theta)&=m_{12}^\star(\theta_1\cap(\theta_2\cup\theta_3))\cdot\frac{m_{12}^\star(\Theta)}{K}=0.005
\end{align*}
\noindent where the normalization constant $K=0.72$ corresponds here to $K=\sum_{Y\in 2^\Theta}m_{12}^\star(Y)$.\\

If one considers now $X=\theta_1\cap\theta_2$ with its belief mass $m_{12}^\star(\theta_1\cap\theta_2)=0.05$, then only $\theta_1$, $\theta_2$ and $\theta_1\cup\theta_2$ enter in the proportional redistribution (version b) because $u(X)=\theta_1\cup\theta_2$ doesn't not carry element $\theta_3$. One then gets for this element $X$ the new set of proportional redistribution factors:
\begin{align*}
k(\theta_1)&=m_{12}^\star(\theta_1\cap\theta_2)\cdot\frac{m_{12}^\star(\theta_1)}{K}
=0.05\cdot \frac{0.20}{0.40}=0.025\\
k(\theta_2)&=m_{12}^\star(\theta_1\cap\theta_2)\cdot\frac{m_{12}^\star(\theta_2)}{K}
 =0.05\cdot \frac{0.17}{0.40}=0.02125\\
k(\theta_1\cup\theta_2)&=m_{12}^\star(\theta_1\cap\theta_2)\cdot\frac{m_{12}^\star(\theta_1\cup\theta_2)}{K}=0.05\cdot \frac{0.03}{0.40}=0.00375
\end{align*}
\noindent where the normalization constant $K=0.40$ corresponds now to the sum $K=m_{12}^\star(\theta_1)+m_{12}^\star(\theta_2)+m_{12}^\star(\theta_1\cup\theta_2)$.\\

The proportional redistribution is done similarly for all other partial conflicting masses. We summarize in the Tables \ref{Table5}-\ref{Table7} all the proportions (rounded at the fifth decimal) of conflicting masses to transfer onto elements of the power set based on this second version of proportional redistribution of  minC.\\

The sum of each column of the Tables \ref{Table5}-\ref{Table7} is transferred onto the mass of the element of power set  it corresponds to get the final result of minC (version b)). By example, $m_{\text{\tiny minC}}^b(\theta_1)$ will be obtained by
\begin{align*}
m_{\text{\tiny minC}}^b(\theta_1)& = m_{12}^\star(\theta_1)+ (0.02500+0.03333+0.01666+0.00834+0.00555)\\
&  = 0.20+0.08888=0.28888
\end{align*}
\noindent
which corresponds to the first value (rounded at the 4th decimal) of the 5th column of Table \ref{Table1}.
All other values of the minC (version b) result of Table \ref{Table1} can be easily verified similarly.
\begin{table}[thb]
\begin{center}
\begin{tabular}{|c||c|c|c|}
\hline
    & $\theta_1$ & $\theta_2$ & $\theta_3$ \\ 
\hline 
$\theta_1\cap\theta_2$ &  0.02500 &  0.02125 &  \\
$\theta_1\cap\theta_3$ &  0.03333 &  &  0.02667\\
$\theta_2\cap\theta_3$ &  & 0.02298 & 0.02162\\
$\theta_1\cap(\theta_2\cup\theta_3)$ &  0.01666 &  0.01417 & 0.01333\\
$\theta_2\cap(\theta_1\cup\theta_3)$ &  0.00834 &  0.00708 & 0.00667\\
$\theta_3\cap(\theta_1\cup\theta_2)$ &  0.00555 &  0.00472 & 0.00444\\
\hline\end{tabular}
\caption{Version b) of minC Proportional conflict redistribution factors}
\label{Table5}
\end{center}
\end{table}
\begin{table}[hhh]
\begin{center}
\begin{tabular}{|c||c|c|}
\hline
    & $\theta_1\cup\theta_2$ & $\theta_1\cup\theta_3$ \\ 
\hline 
$\theta_1\cap\theta_2$ &  0.00375 &  \\
$\theta_1\cap\theta_3$ &  & 0.01000  \\
$\theta_2\cap\theta_3$ &  &  \\
$\theta_1\cap(\theta_2\cup\theta_3)$ & 0.00250 & 0.00500 \\
$\theta_2\cap(\theta_1\cup\theta_3)$ & 0.00125 & 0.00250 \\
$\theta_3\cap(\theta_1\cup\theta_2)$ & 0.00084 & 0.00167\\
\hline\end{tabular}
\caption{Version b) of minC Proportional conflict redistribution factors (continued)}
\label{Table6}
\end{center}
\end{table}
\begin{table}[hhh]
\begin{center}
\begin{tabular}{|c||c|c|}
\hline
    &  $\theta_2\cup\theta_3$ & $\theta_1\cup\theta_2\cup\theta_3$\\ 
\hline 
$\theta_1\cap\theta_2$ &  & \\
$\theta_1\cap\theta_3$ &  & \\
$\theta_2\cap\theta_3$ &  0.00540 &\\
$\theta_1\cap(\theta_2\cup\theta_3)$ & 0.00333 & 0.00500\\
$\theta_2\cap(\theta_1\cup\theta_3)$ & 0.00166 & 0.00250\\
$\theta_3\cap(\theta_1\cup\theta_2)$ & 0.00111&  0.00167\\
\hline\end{tabular}
\caption{Version b) of minC Proportional conflict redistribution factors (continued)}
\label{Table7}
\end{center}
\end{table}
\end{itemize}

\section{Principle of the PCR rules}

Let's $\Theta= \{\theta_1, \theta_2,\ldots, \theta_n\}$ be the frame of the fusion problem under consideration and two belief assignments $m_1, m_2 : G \rightarrow
[0, 1]$ such that $\sum_{X\in G} m_i(X)=1$, $i=1,2$.  The general principle of the Proportional Conflict Redistribution Rules (PCR for short) is:
\begin{itemize}
\item apply the conjunctive rule \eqref{eq:ConjPS}  or \eqref{eq:ConjHPS}  depending on theory, i.e. $G$ can be either $2^\Theta$ or $D^\Theta$,
\item calculate the total or partial conflicting masses,
\item then redistribute the conflicting mass (total or partial) proportionally on non-empty sets involved in the model according to all integrity constraints.
\end{itemize}
The way the conflicting mass is redistributed yields to five versions of PCR, denoted PCR1, PCR2, \ldots , PCR5 as it will be shown in the sequel. The PCR combination rules work for any degree of conflict $k_{12} \in [0, 1]$ or $k_{12\ldots s} \in [0, 1]$, for any DSm models (Shafer's model, free DSm model or any hybrid DSm model). PCR rules work both in DST and DSmT frameworks and for static or dynamical fusion problematics.
The sophistication/complexity (but correctness) of proportional conflict redistribution increases from the first PCR1 rule up to the last rule PCR5.
The development of different PCR rules presented here comes from the fact that the first initial PCR rule developed (PCR1) does not preserve the neutral impact of VBA. All other improved rules PCR2-PCR5 preserve the commutativity, the neutral impact of VBA and propose, upon to our opinion, a more and more exact solution for the conflict management to satisfy as best as possible the condition 1 (in section 1) that any satisfactory combination rule must tend to. The general proof for the neutrality of VBA within PCR2, PCR3, PCR4 and PCR5 rules is given in section \ref{SectionPCR5} and some numerical examples are given in the section related with the presentation of each rule.

\section{The PCR1 rule}

\subsection{The PCR1 formula}

PCR1 is the simplest and the easiest version of proportional conflict redistribution for combination.
PCR1 is described in details in \cite{Smarandache_Dezert_2004c}. The basic idea for PCR1 is only to compute the total conflicting mass $k_{12}$ (not worrying about the partial conflicting masses). The total conflicting mass is then distributed to {\it{all non-empty sets}} proportionally with respect to their corresponding non-empty column sum of the associated mass matrix. The PCR1 is defined $\forall (X\neq\emptyset)\in G$ by:
\begin{itemize}
\item For the the combination of $s=2$ sources
\begin{equation}
m_{PCR1}(X)=[\sum_{\substack{X_1,X_2\in G \\ X_1\cap X_2=X}}m_{1}(X_1)m_{2}(X_2)]
 + \frac{c_{12}(X)}{d_{12}}\cdot k_{12}
\end{equation}
\noindent
where $c_{12}(X)$ is the non-zero sum of the column of $X$ in the mass matrix 
$\mathbf{M}=\begin{bmatrix}
\mathbf{m}_1\\
\mathbf{m}_2
\end{bmatrix}
$ 
(where $\mathbf{m}_i$ for $i=1,2$ is the row vector of belief assignments committed by the source $i$ to elements of $G$), i.e. $c_{12}(X)=
m_1(X)+m_2(X) \neq 0$, $k_{12}$ is the total conflicting mass, and $d_{12}$ is the sum of all non-zero column sums of all non-empty sets (in many cases $d_{12} = 2$, but in some degenerate cases it can be less) (see \cite{Smarandache_Dezert_2004c}).\\
\item For the the combination of $s\geq 2$ sources
\begin{equation}
m_{PCR1}(X)=[\sum_{\substack{X_1,X_2,\ldots,X_s\in G \\ X_1\cap X_2\cap\ldots\cap X_s=X}}\prod_{i=1}^{s}m_{i}(X_i)]
+ \frac{c_{12\ldots s}(X)}{d_{12\ldots s}}\cdot k_{12\ldots s}
\end{equation}

\noindent
where $c_{12\ldots s}(X)$ is the non-zero sum of the column of $X$ in the mass matrix, i.e. $c_{12\ldots s}(X)=
m_1(X)+m_2(X)+\ldots + m_s(X) \neq 0$, $k_{12\ldots s}$ is the total conflicting mass, and $d_{12\ldots s}$ is the sum of all non-zero column sums of all non-empty sets (in many cases $d_{12\ldots s} = s$, but in some degenerate cases it can be less).
\end{itemize}

PCR1 is an alternative combination rule to WAO (Weighted Average Operator) proposed by J{\o}sang, Daniel and Vannoorenberghe in \cite{Josang_2003}. Both are particular cases of WO (The Weighted Operator) because the conflicting mass is redistributed with respect to some weighting factors. In the PCR1, the proportionalization is done for each non-empty set with respect to the non-zero sum of its corresponding mass matrix - instead of its mass column average as in WAO. But, PCR1 extends WAO, since PCR1 works also for the degenerate cases when all column sums of all non-empty
sets are zero because in such cases, the conflicting mass is transferred to the non-empty disjunctive form of all non-empty sets together; when this disjunctive form happens to be empty, then either the problem degenerates truly to a void problem and thus all conflicting mass is transferred onto the empty set,  or we can assume (if one has enough reason to justify such assumption) that  the frame of discernment might contain new unknown hypotheses all summarized by $\theta_0$ and under this assumption all conflicting mass is transferred onto the unknown possible $\theta_0$.\\

A nice feature of PCR1 rule, is that it works in all cases (degenerate and non degenerate). PCR1 corresponds to a specific choice of proportionality coefficients in the infinite continuum family\footnote{pointed out independently by Inagaki in 1991 and Lef\`evre,
Colot and Vannoorenberghe in 2002.} of possible rules of combination involving conjunctive consensus operator. The PCR1 on the power set and for non-degenerate cases gives the same results as WAO (as Philippe Smets pointed out); yet, for the storage proposal in a dynamic fusion when the associativity is needed, for PCR1 is needed to store only the last sum of masses, besides the previous conjunctive ruleÕs result, while in WAO it is in addition needed to store the number of the
steps (see \cite{Smarandache_Dezert_2004c} for details) Ð and both rules become quasi-associative. 
In addition to WAO, we propose a general
formula for PCR1 (WAO for non-degenerate cases).\\

Unfortunately, a severe limitation of PCR1 (as for WAO) is the non-preservation of the neutral impact of the VBA as shown in \cite{Smarandache_Dezert_2004c}. In other words, for $s\geq 1$, one gets for $m_1(.)\neq m_v(.)$, \ldots, $m_s(.)\neq m_v(.)$:
\begin{equation*}
m_{PCR1}(.)=[m_1\oplus\ldots m_s\oplus m_v](.) \neq [m_1\oplus\ldots m_s](.)
\end{equation*}
For the cases of the combination of only one non-vacuous belief assignment $m_1(.)$ with the vacuous belief assignment $m_v(.)$ where $m_1(.)$ has mass assigned to an empty element, say
$m_1(\emptyset)>0$ as in Smets' TBM, or as in DSmT dynamic
fusion where one finds out that a previous non-empty element
$A$, whose mass $m_1(A)>0$, becomes empty after a certain
time, then this mass of an empty set has to be transferred to other elements using PCR1, but for such case $[m_1\oplus m_v](.)$ is different from $m_1(.)$. This severe drawback of WAO and PCR1 forces us to develop the next PCR rules satisfying the neutrality property of VBA with better redistributions of the conflicting information.

\subsection{Example for PCR1 (degenerate case)}

For non degenerate cases with Shafer's model, PCR1 and WAO provide the same results. So it is interesting to focus the reader's attention on the difference between PCR1 and WAO in a simple degenerate case corresponding to a dynamic fusion problem. Let's take the following example showing the restriction of applicability of {\it{static}}-WAO\footnote{{\it{static}}-WAO stands for the WAO rule proposed in \cite{Lefevre_2002,Josang_2003} based on Shafer's model for the implicit static fusion case (i.e. $\Theta$ remains invariant with time), while {\it{dynamic}}-WAO corresponds to the Daniel's improved version of WAO using \eqref{eq:wm_WAODyn}.}. As example, let's consider three different suspects $A$, $ B$ and $C$ in a criminal investigation (i.e. $\Theta=\{A,B,C\}$) and the two following simple Bayesian witnesses reports
$$m_1(A)=0.3 \quad m_1(B)=0.4\quad m_1(C)=0.3$$
$$m_2(A)=0.5\quad m_2(B)=0.1\quad m_2(C)=0.4$$
\noindent The conjunctive consensus is 
$$m_{12}(A)=0.15\quad m_{12}(B)=0.04\quad m_{12}(C)=0.12$$
\noindent
with the conflicting mass $k_{12}=0.69$.
Now let's assume that a little bit later, one learns that $B=\emptyset$ because the second suspect brings a strong alibi, then the initial consensus on $B$ (i.e. $m_{12}(B)=0.04$) must enter now in the new conflicting mass $k_{12}'=0.69+0.04=0.73$ since $B=\emptyset$. Applying the PCR1 formula, one gets now:
\begin{align*}
m_{PCR1|12}(B)& = 0\\
m_{PCR1|12}(A) & = 0.15 + \frac{0.8}{0.8+0.7}\cdot 0.73 = 0.5393\\
m_{PCR1|12}(C) & = 0.12 + \frac{0.7}{0.8+0.7}\cdot 0.73 = 0.4607
\end{align*}
\noindent
Let's remind (see section 4.3) that in this case, the {\it{static}}-WAO provides
$$m_{WAO|12}(B) = 0 \qquad m_{WAO|12}(A) = 0.4420 \qquad m_{WAO|12}(C) = 0.3755$$
We can verify easily that $m_{PCR1|12}(A)+m_{PCR1|12}(B)+m_{PCR1|12}(C)=1$ while
$m_{WAO|12}(A)+m_{WAO|12}(B)+m_{WAO|12}(C)=0.8175 < 1$.
This example shows clearly the difference between PCR1 and {\it{static}}-WAO originally proposed in \cite{Lefevre_2002,Josang_2003} and the ability of PCR1 to deal with degenerate/dynamic cases contrariwise to original WAO. The improved {\it{dynamic}}-WAO version suggested  by Daniel coincides with PCR1.

\section{The PCR2 rule}

\subsection{The PCR2 formula}

In PCR2, the total conflicting mass $k_{12}$ is
distributed only to {\it{the non-empty sets involved in the conflict}} (not to all non-empty sets) and taken the canonical form of the conflict proportionally with respect to their corresponding non-empty column sum. The redistribution is then more exact (accurate) than in PCR1 and WAO. A nice feature of PCR2 is the preservation of the neutral impact of the  VBA and of course its ability to deal with all cases/models.\\

A non-empty set $X_1\in G$ is considered {\it{involved in the conflict}} if there exists another set 
$X_2\in G$ which is neither included in $X_1$ nor includes $X_1$ such that $X_1\cap X_2=\emptyset$ and $m_{12}(X_1\cap X_2)>0$. This definition can be generalized for $s \geq 2$ sources.
\begin{itemize}
\item
The PCR2 formula for two sources ($s=2$) is  $\forall (X\neq\emptyset)\in G$, 
\begin{equation}
m_{PCR2}(X)=[\sum_{\substack{X_1,X_2\in G \\ X_1\cap X_2=X}}m_{1}(X_1)m_{2}(X_2)]
 + \mathcal{C}(X)\frac{c_{12}(X)}{e_{12}} \cdot k_{12} 
\end{equation}
\noindent
where
\begin{equation*}
\mathcal{C}(X)=
\begin{cases}
1, \quad \text{if $X$ involved in the conflict,}\\
0,\quad\text{otherwise;}
\end{cases}
\end{equation*}

\noindent
and where $c_{12}(X)$ is the non-zero sum of the column of $X$ in the mass matrix, i.e. $c_{12}(X)=
m_1(X)+m_2(X) \neq 0$, $k_{12}$ is the total conflicting mass, and $e_{12}$ is the sum of all non-zero column sums of all non-empty sets {\it{only}} involved in the conflict (resulting from the conjunctive normal form of their intersection after using the conjunctive rule). In many cases $e_{12} = 2$, but in some degenerate cases it can be less.\\

\item For the the combination of $s\geq 2$ sources, the previous PCR2 formula can be easily generalized as follows $\forall (X\neq\emptyset)\in G$:
\begin{equation}
m_{PCR2}(X)=[\sum_{\substack{X_1,X_2,\ldots,X_s\in G \\ X_1\cap X_2\cap\ldots\cap X_s=X}}\prod_{i=1}^{s}m_{i}(X_i)] + \mathcal{C}(X)\frac{c_{12\ldots s}(X)}{e_{12\ldots s}}\cdot k_{12\ldots s}
\end{equation}
\noindent
where
\begin{equation*}
\mathcal{C}(X)=
\begin{cases}
1, \quad \text{if $X$ involved in the conflict,}\\
0,\quad\text{otherwise;}
\end{cases}
\end{equation*}
\noindent
and $c_{12\ldots s}(X)$ is the non-zero sum of the column of $X$ in the mass matrix, i.e. $c_{12\ldots s}(X)=
m_1(X)+m_2(X)+\ldots + m_s(X) \neq 0$, $k_{12\ldots s}$ is the total conflicting mass, and $e_{12\ldots s}$ is the sum of all non-zero column sums of all non-empty sets involved in the conflict (in many cases $e_{12\ldots s} = s$, but in some degenerate cases it can be less).
\end{itemize}

In the degenerate case when all column sums of all non-empty sets involved in the conflict
are zero, then the conflicting mass is transferred to the non-empty disjunctive form of all
sets together which were involved in the conflict together. But if this disjunctive form
happens to be empty, then the problem reduces to a degenerate void problem and thus all conflicting mass is transferred to the empty set or we can assume (if one has enough reason to justify such assumption) that  the frame of discernment might contain new unknown hypotheses all summarized by $\theta_0$ and under this assumption all conflicting mass is transferred onto the unknown possible $\theta_0$.

\subsection{Example for PCR2 versus PCR1}

 LetÕs have the frame of discernment $\Theta=\{A, B\}$, Shafer's model (i.e. all intersections empty),
and the following two bbas:
$$m_1(A)=0.7 \quad m_1(B)=0.1 \quad m_1(A\cup B)=0.2$$
$$m_2(A)=0.5 \quad m_2(B)=0.4 \quad m_2(A\cup B)=0.1$$
The sums of columns of the mass matrix are
$$c_{12}(A)=1.2 \quad c_{12}(B)=0.5 \quad c_{12}(A\cup B)=0.3$$
\noindent
Then the conjunctive consensus yields 
$$m_{12}(A)=0.52\quad m_{12}(B)=0.13\quad m_{12}(A\cup B)=0.02$$
\noindent
with the total conflict $k_{12}=m_{12}(A\cap B)  = 0.33$.\\
\noindent

\begin{itemize}
\item
Applying the PCR1 rule yields ($d_{12}=1.2+0.5+0.3=2$):
\begin{align*}
m_{PCR1|12}(A)&=m_{12}(A) + \frac{c_{12}(A)}{d_{12}}\cdot k_{12}=0.52 + \frac{1.2}{2}\cdot 0.33 =0.7180\\
m_{PCR1|12}(B)&=m_{12}(B) +\frac{c_{12}(B)}{d_{12}}\cdot k_{12} =0.13+ \frac{0.5}{2}\cdot 0.33 =0.2125\\
m_{PCR1|12}(A\cup B)&=m_{12}(A\cup B)+\frac{c_{12}(A\cup B)}{d_{12}}\cdot k_{12}
=0.02+ \frac{0.3}{2}\cdot 0.33 =0.0695
\end{align*}
\end{itemize}
\begin{itemize}
\item
While applying the PCR2 rule yields ($e_{12}=1.2+0.5=1.7$):
\begin{align*}
m_{PCR2}(A)&=m_{12}(A) + \frac{c_{12}(A)}{e_{12}}\cdot k_{12}=0.52 + \frac{1.2}{1.7}\cdot 0.33 =0.752941\\
m_{PCR2}(B)&=m_{12}(B) +\frac{c_{12}(B)}{e_{12}}\cdot k_{12}=0.12 + \frac{0.5}{1.7}\cdot 0.33 =0.227059\\
m_{PCR2}(A\cup B)&=m_{12}(A\cup B)=0.02
\end{align*}
\end{itemize}

\subsection{Example of neutral impact of VBA for PCR2}

Let's keep the previous example and introduce now a third but totally ignorant source $m_v(.)$ and examine the result of the combination of the 3 sources with PCR2. So, let's start with
$$m_1(A)=0.7 \quad m_1(B)=0.1 \quad m_1(A\cup B)=0.2$$
$$m_2(A)=0.5 \quad m_2(B)=0.4 \quad m_2(A\cup B)=0.1$$
$$m_v(A)=0 .0\quad m_v(B)=0.0 \quad m_v(A\cup B)=1.0$$
The sums of columns of the mass matrix are
$$c_{12v}(A)=1.2 \quad c_{12v}(B)=0.5 \quad c_{12v}(A\cup B)=1.3$$
\noindent
Then the conjunctive consensus yields 
$$m_{12v}(A)=0.52\quad m_{12v}(B)=0.13\quad m_{12v}(A\cup B)=0.02$$
\noindent
with the total conflict $k_{12v}=m_{12v}(A\cap B)  = 0.33$. We get naturally $m_{12v}(.)=m_{12}(.)$ because the vacuous belief assignment $m_v(.)$ has no impact in the conjunctive consensus.\\

Applying the PCR2 rule yields:
\begin{align*}
m_{PCR2|12v}(A)&=m_{12v}(A) +\frac{c_{12v}(A)}{e_{12v}}\cdot k_{12v}=0.52 + \frac{1.2}{1.2+0.5}\cdot 0.33 =0.752941\\
m_{PCR2|12v}(B)&=m_{12v}(B) + \frac{c_{12v}(B)}{e_{12v}}\cdot k_{12v}=0.52 + \frac{0.5}{1.2+0.5}\cdot 0.33 =0.227059\\
m_{PCR2|12v}(A\cup B)&=m_{12v}(A\cup B)=0.02
\end{align*}
In this example one sees that the neutrality property of VBA is effectively well satisfied since
$$m_{PCR2|12v}(.)=m_{PCR2|12}(.)$$

\noindent
A general proof for neutrality of VBA within PCR2 is given in section \ref{SectionPCR5}.\\

\section{The PCR3 rule}
\subsection{Principle of  PCR3}

In PCR3, one transfers {\it{partial conflicting masses}}, instead of the total conflicting mass, to non-empty sets involved in partial conflict (taken the canonical form of each partial conflict). If an intersection is empty, say $A\cap B = \emptyset$, then the mass $m(A\cap B)$ of the partial conflict is transferred to the non-empty sets $A$ and $B$ proportionally with respect to the non-zero sum of masses assigned to $A$ and respectively to $B$ by the bbas $m_1(.)$ and  $m_2(.)$. The PCR3 rule works if at least one set between $A$ and $B$ is
non-empty and its column sum is non-zero.\\

When both sets $A$ and $B$ are empty, or both corresponding column sums of the mass matrix are zero, or only one set is non-empty and its column sum is zero, then the mass $m(A\cap B)$ is transferred to the non-empty disjunctive form $u(A)\cup u(B)$ defined in \eqref{eq:u}; if this disjunctive form is empty then $m(A\cap B)$ is transferred to the non-empty total ignorance; but if even the total ignorance is empty then either the problem degenerates truly to a void problem and thus all conflicting mass is transferred onto the empty set,  or we can assume (if one has enough reason to justify such assumption) that  the frame of discernment might contain new unknown hypotheses all summarized by $\theta_0$ and under this assumption all conflicting mass is transferred onto the unknown possible $\theta_0$.\\

If another intersection, say $A\cap C\cap D = \emptyset$, then again the mass $m(A\cap C\cap D)>0$ is transferred to the non-empty sets $A$, $C$, and $D$ proportionally with respect to the non-zero sum of masses assigned to $A$, $C$, and respectively $D$ by the sources; if all
three sets $A$, $C$, $D$ are empty or the sets which are non-empty have their corresponding
column sums equal to zero, then the mass $m(A\cap C\cap D)$ is transferred to the non-empty
disjunctive form $u(A)\cup u(C)\cup u(D)$; if this disjunctive form is empty then the mass
$m(A\cap C\cap D)$ is transferred to the non-empty total ignorance; but if even the total ignorance is empty (a completely degenerate void case)  all conflicting mass is transferred onto the empty set (which means that the problem is truly void), or (if we prefer to adopt an optimistic point of view) all conflicting mass is transferred onto a new unknown extra and closure element $\theta_0$ representing all missing hypotheses of the frame $\Theta$.\\

 The {\it{disjunctive form}} is defined\footnote{These relationships can be generalized for any number of sets.} as \cite{DSmTBook_2004a}: 
 \begin{equation}
\begin{cases}
 u(X) = X\,\text{ if}\, X \,\text{is a singleton}\\
 u(X\cup Y) = u(X)\cup u(Y)\\
u(X\cap Y) = u(X)\cup u(Y)
 \end{cases}
\label{eq:u}
\end{equation}
 \noindent
 
 \clearpage
 \newpage
 
 \subsection{The PCR3 formula}

\begin{itemize}

\item 
For the combination of two bbas, the PCR3 formula is given by: $\forall (X\neq\emptyset)\in G$,
\begin{align}
&m_{PCR3}(X)=[\sum_{\substack{X_1,X_2\in G \\ X_1\cap X_2=X}}m_{1}(X_1)m_{2}(X_2)]+ [c_{12}(X)\cdot 
 \sum_{\substack{Y\in G \\ c(Y\cap X)=\emptyset}}\frac{m_{1}(Y)m_{2}(X)+m_{1}(X)m_{2}(Y)}{c_{12}(X)+c_{12}(Y)}]\nonumber\\
 &+[\sum_{\substack{X_1,X_2\in (G\setminus\{X\})\cap\boldsymbol{\emptyset}\\ c(X_1\cap X_2)=\emptyset\\ u(X_1)\cup u(X_2)=X}}[m_{1}(X_1)m_{2}(X_2) + m_{1}(X_2)m_{2}(X_1)]]\nonumber\\
 &+ [\phi_\Theta(X)
 \sum_{\substack{X_1,X_2\in (G\setminus\{X\})\cap\boldsymbol{\emptyset} \\ c(X_1\cap X_2)=\emptyset\\ u(X_1)= u(X_2)=\emptyset}}[m_{1}(X_1)m_{2}(X_2) + m_{1}(X_2)m_{2}(X_1)]]
  \label{eq:PCR3}
 \end{align}

\noindent 
where $c(\alpha)$ is the conjunctive normal (i.e. canonical) form of $\alpha$, where $\alpha$ is in $G$, $c_{12}(X_i)$ ($X_i\in G$) is the non-zero sum of the mass matrix column corresponding to the set $X_i$, i.e. $c_{12}(X_i) = m_1(X_i) + m_2(X_i) \neq 0$, and where $\phi_\Theta(.)$ is the characteristic function of the total ignorance (assuming $\mid\Theta\mid=n$) defined by
\begin{equation}
\begin{cases}
\phi_\Theta(X)=1\, \text{if}\, X=\theta_1\cup\theta_2\cup\ldots\cup\theta_n \, \text{(total ignorance)}\\
\phi_\Theta(X)=0 \quad \text{otherwise}
\end{cases}
\label{eq:CFTI}
\end{equation}

\item 
For the fusion of $s \geq  2$ bbas, one extends the above procedure to formulas \eqref{eq:u} and  \eqref{eq:PCR3} to more general ones. One then gets the following PCR3 general formula. 
Let $G=\{X_1,\ldots,X_n\}\neq \emptyset$ ($G$ being either the power-set or hyper-power set depending on the model we want to deal with), $n\geq 2$, $\forall X\neq\emptyset$, $X\in G$, one has:
\begin{align}
m_{PCR3}(X)&=m_{12\ldots s}(X)+ c_{12\ldots s}(X)\cdot\sum_{k=1}^{s-1} 
S_1^{PCR3}(X,k)+\sum_{k=1}^{s}S_2^{PCR3}(X,k)+ \phi_\Theta(X)
\sum_{k=1}^s S_3^{PCR3}(X,k)
  \label{eq:PCR3s}
 \end{align}

\noindent  For convenience, the following notation is used

$$m_{12\ldots s}(X)=\sum_{\substack{X_1,\ldots,X_s\in G \\ X_1\cap \ldots\cap X_s=X}}\prod_{k=1}^s m_{k}(X_k)$$
 $$m_{12\ldots s}(\bigcap_{j=1}^{k} X_{i_j})=m_{12\ldots s}(X_{i_1}\cap \ldots \cap X_{i_k})$$

$$S_1^{PCR3}(X,k)\triangleq \sum_{\substack{X_{i_1},\ldots,X_{i_k}\in G\setminus\{X\}\\
 \{i_1,\ldots,i_k\} \in \mathcal{P}^k(\{1,2,\ldots,n\})
 \\ c(X\cap X_{i_1}\cap\ldots\cap X_{i_k})=\emptyset}}
 R_k^{i_1,\ldots,i_k}(X)$$

\noindent
with

$$ R_k^{i_1,\ldots,i_k}(X)
 \triangleq \frac{m_{12\ldots s}(X\cap X_{i_1}\cap\ldots\cap X_{i_k})}{c_{12\ldots s}(X)+\sum_{j=1}^k c_{12\ldots s}(X_{i_j})}$$

\noindent
and
$$S_2^{PCR3}(X,k)\triangleq\sum_{\substack{X_{i_1},\ldots, X_{i_k}\in (G\setminus\{X\})\cap\boldsymbol{\emptyset} \\
\{i_1,\ldots,i_k\} \in \mathcal{P}^k(\{1,2,\ldots,n\})
 \\ c(X_{i_1}\cap \ldots \cap X_{i_k})=\emptyset\\ u(X_{i_1})\cup \ldots \cup u(X_{i_k})=X}}
 m_{12\ldots s}(\bigcap_{j=1}^{k} X_{i_j})$$

$$S_3^{PCR3}(X,k)\triangleq \sum_{\substack{X_{i_1},\ldots, X_{i_k}\in (G\setminus\{X\})\cap\boldsymbol{\emptyset}\\
\{i_1,\ldots,i_k\} \in \mathcal{P}^k(\{1,2,\ldots,n\})
 \\ c(X_{i_1}\cap \ldots \cap X_{i_k})=\emptyset\\ u(X_{i_1})= \ldots = u(X_{i_k})=\emptyset}}
 m_{12\ldots s}(\bigcap_{j=1}^{k} X_{i_j})$$

\noindent
where $\boldsymbol{\emptyset}$ is the set of elements (if any) which have been forced to be empty by the integrity constraints of the model of the problem (in case of dynamic fusion) and ($\mathcal{P}^{k}(\{1,2,\ldots,n\})$ is the set of all subsets ok $k$ elements from $\{1,2,\ldots,n\}$ (permutations of $n$ elements taken by $k$), the order of elements doesn't count.\\

The sum $\sum_{k=1}^{s}S_2^{PCR3}(X,k)$ in \eqref{eq:PCR3s} is for cases when $X_{i_1}$,\ldots, $X_{i_k}$ become empty in dynamic fusion; their intersection mass is transferred to their disjunctive form: $u(X_{i_1})\cup \ldots \cup u(X_{i_k})\neq\emptyset$.\\

The sum $\sum_{k=1}^s S_3^{PCR3}(X,k)$ in \eqref{eq:PCR3s} is for degenerate cases, i.e. when $X_{i_1}$,\ldots, $X_{i_k}$ and their disjunctive form become empty in dynamic fusion; their intersection mass is transferred to the total ignorance.
\end{itemize}
\noindent
PCR3 preserves the neutral impact of the VBA and works for any cases/models.

\subsection{Example for PCR3}

Let's have the frame of discernment $\Theta=\{A, B, C\}$, Shafer's model (i.e. all intersections
empty), and the 2 following Bayesian bbas
$$m_1(A)=0.6 \quad m_1(B)=0.3 \quad m_1(C)=0.1$$
$$m_2(A)=0.4 \quad m_2(B)=0.4 \quad m_2(C)=0.2$$
\noindent
The sums of columns of the mass matrix are
$$c_{12}(A)=1.0 \quad c_{12}(B)=0.7 \quad c_{12}(C)=0.3$$
\noindent
Then the conjunctive consensus yields 
$$m_{12}(A)=0.24\quad m_{12}(B)=0.12\quad m_{12}(C)=0.02$$
\noindent
with the total conflict $k_{12}=m_{12}(A\cap B) + m_{12}(A\cap C) + m_{12}(B\cap C) = 0.36 + 0.16 + 0.10 = 0.62$, which is a sum of factors.\\
\noindent

Applying the PCR3 rule yields for this very simple (Bayesian) case:
\begin{align*}
m_{PCR3|12}(A)&=m_{12}(A) +c_{12}(A)\cdot \frac{m_1(B)m_2(A)+m_1(A)m_2(B)}{c_{12}(A)+c_{12}(B)}+c_{12}(A)\cdot \frac{m_1(C)m_2(A)+m_1(A)m_2(C)}{c_{12}(A)+c_{12}(C)}\\
&=0.24 + 1\cdot \frac{0.3\cdot 0.4 + 0.6\cdot 0.4}{1 + 0.7} +  1\cdot \frac{0.1\cdot 0.4 + 0.6\cdot 0.2}{1 + 0.3}
=0.574842
\end{align*}
\begin{align*}
m_{PCR3|12}(B)&=m_{12}(B) +c_{12}(B)\cdot \frac{m_1(A)m_2(B)+m_1(B)m_2(A)}{c_{12}(B)+c_{12}(A)}+c_{12}(B)\cdot \frac{m_1(C)m_2(B)+m_1(B)m_2(C)}{c_{12}(B)+c_{12}(C)}\\
 &=0.12 + 0.7\cdot \frac{0.6\cdot 0.4 + 0.3\cdot 0.4}{0.7+1} +  0.7\cdot \frac{0.1\cdot 0.4 + 0.3\cdot 0.2}{0.7+0.3}
=0.338235
\end{align*}
\begin{align*}
m_{PCR3|12}(C)&=m_{12}(C) 
+c_{12}(C)\cdot \frac{m_1(C)m_2(A)+m_1(A)m_2(C)}{c_{12}(C)+c_{12}(A)}
+c_{12}(C)\cdot \frac{m_1(C)m_2(B)+m_1(B)m_2(C)}{c_{12}(C)+c_{12}(B)}\\
&=0.02 + 0.3\cdot \frac{0.1\cdot 0.4 + 0.6\cdot 0.2}{0.3+1} 
+  0.3\cdot \frac{0.1\cdot 0.4 + 0.2\cdot 0.3}{0.3+0.7}
 =0.086923
\end{align*}

Note that in this simple case, the two last sums involved in formula \eqref{eq:PCR3} are equal to zero because here there doesn't exist positive mass products $m_{1}(X_1)m_{2}(X_2)$ to compute for any $X\in 2^\Theta$, $X_1,X_2\in 2^\Theta \setminus\{X\}$ such that $X_1\cap X_2=\emptyset$ and $u(X_1)\cup u(X_2)=X$, neither for $X_1\cap X_2=\emptyset$ and $u(X_1)= u(X_2)=\emptyset$.\\

In this example, PCR3 provides a result different from PCR1 and PCR2 (PCR2 provides same result as PCR1) since
\begin{align*}
& m_{PCR1}(A) = 0.24 + \frac{1}{1+0.7+0.3}\cdot 0.62 = 0.550\\
& m_{PCR1}(B) = 0.12 + \frac{0.7}{1+0.7+0.3}\cdot 0.62  = 0.337\\
& m_{PCR1}(C) = 0.02 + \frac{0.3}{1+0.7+0.3}\cdot 0.62 = 0.113
\end{align*}

\subsection{Example of neutral impact of VBA for PCR3}

Let's keep the previous example and introduce now a third but totally ignorant source $m_v(.)$ and examine the result of the combination of the 3 sources with PCR3. $\Theta$ denotes here for notation convenience $A\cup B\cup C$. So, Let's start with
\begin{align*}
m_1(A)=0.6 & \quad m_1(B)=0.3 &  m_1(C)=0.1 & \\
m_2(A)=0.4 & \quad m_2(B)=0.4 & m_2(C)=0.2 &\\
m_v(A)=0 .0 & \quad m_v(B)=0.0 & m_v(C)=0.0 &\ m_v(\Theta)=1
\end{align*}

\noindent
The sums of columns of the mass matrix are
$$c_{12v}(A)=1, \  c_{12v}(B)=0.7, \  c_{12v}(C)=0.3, \   c_{12v}(\Theta)=1$$
\noindent
The conjunctive consensus yields 
$$m_{12v}(A)=0.24\quad m_{12v}(B)=0.12\quad m_{12v}(C)=0.02$$
\noindent
with the total conflict $k_{12v}=m_{12v}(A\cap B) + m_{12v}(A\cap C) + m_{12v}(B\cap C)= 0.36 + 0.16 + 0.10 = 0.62$, which is a sum of factors. We get naturally $m_{12v}(.)=m_{12}(.)$ because the vacuous belief assignment $m_v(.)$ has no impact on the conjunctive consensus.\\

\noindent
Applying the PCR3 rule yields for this case
\begin{align*}
m_{PCR3|12v}(A)=&m_{12v}(A) \\
&+ c_{12v}(A)\cdot [ \frac{m_1(B)m_2(A)m_v(\Theta)}{c_{12v}(A)+c_{12v}(B)}+ \frac{m_1(A)m_2(B)m_v(\Theta)}{c_{12v}(A)+c_{12v}(B)}]\\
&+c_{12v}(A)\cdot [\frac{m_1(C)m_2(A)m_v(\Theta)}{c_{12v}(A)+c_{12v}(C)}
+ \frac{m_1(A)m_2(C)m_v(\Theta)}{c_{12v}(A)+c_{12v}(C)}]\\
 =&0.24 + 1\cdot \frac{0.3\cdot 0.4 \cdot 1+ 0.6\cdot 0.4\cdot 1}{1 + 0.7} 
+  1\cdot \frac{0.1\cdot 0.4 \cdot 1+ 0.6\cdot 0.2\cdot 1}{1 + 0.3}\\
=&0.574842=m_{PCR3|12}(A)
\end{align*}
Similarly, one obtains
\begin{align*}
m_{PCR3|12v}(B)=&0.12 + 0.7\cdot \frac{0.6\cdot 0.4 \cdot 1+ 0.3\cdot 0.4\cdot 1}{0.7+1} 
+  0.7\cdot \frac{0.1\cdot 0.4\cdot 1 + 0.3\cdot 0.2\cdot 1}{0.7+0.3}\\
=&0.338235=m_{PCR3|12}(B)
\end{align*}
\begin{align*}
m_{PCR3|12v}(C)= &0.02 + 0.3\cdot \frac{0.1\cdot 0.4\cdot 1 + 0.6\cdot 0.2\cdot 1}{0.3+1} 
+  0.3\cdot \frac{0.1\cdot 0.4\cdot 1 + 0.2\cdot 0.3\cdot 1}{0.3+0.7} \\
=& 0.086923=m_{PCR3|12}(C)
\end{align*}

\noindent
In this example one sees that the neutrality property of VBA is effectively well satisfied by PCR3 rule since
$$m_{PCR3|12v}(.)=m_{PCR3|12}(.)$$
\noindent A general proof for neutrality of VBA within PCR3 is given in section \ref{SectionPCR5}.

\section{The PCR4 rule}

\subsection{Principle of PCR4}

PCR4 redistributes the partial conflicting mass to the elements involved in the partial conflict, considering the canonical form of the partial conflict. PCR4 is an improvement of previous PCR rules but also of Milan Daniel's minC operator \cite{DSmTBook_2004a}.
Daniel uses the proportionalization with respect to the results of  the conjunctive rule, but not with respect to the masses assigned to each set by the sources of information as done in PCR1-3  and  also as in the most effective PCR5 rule explicated in the next section.
Actually, PCR4 also uses the proportionalization with respect to the results of the conjunctive rule, but with PCR4 the conflicting mass $m_{12}(A\cap B)$ when $A\cap B=\emptyset$ is distributed to $A$ and $B$ only because only $A$ and $B$ were involved in the conflict  ($A\cup B$ was not involved in the conflict since $m_{12}(A\cap B) = m_1(A)m_2(B)+m_2(A)m_1(B)$), while minC redistributes $m_{12}(A\cap B)$ to $A$, $B$, {\it{and}} $A\cup B$ in both of its versions a) and b) (see section 5 and \cite{DSmTBook_2004a} for details). Also, for the mixed elements such as $C\cap(A\cup B)=\emptyset$, the mass $m( C\cap (A\cup B) )$ is redistributed to $C$, $A\cup B$, $A\cup B\cup C$ in minC version a), and worse in minC version b) to $A$, $B$, $C$, $A\cup B$, $A\cup C$, $B\cup C$ and $A\cup B\cup C$ (see example in section 5). PCR4 rule improves this and redistributes the mass $m( C\cap (A\cup B) )$ to $C$ and $A\cup B$ {\it{only}}, since only them were involved in the conflict: i.e. $m_{12}(C\cap (A\cup B))= m_1(C)m_2(A\cup B)+m_2(C)m_1(A\cup B$), clearly the other elements $A$, $B$, $A\cup B\cup C$ that get some mass in minC were not involved in the conflict $C\cap (A\cup B)$. If at least one conjunctive rule result is null, then the partial conflicting mass which involved this set is redistributed proportionally to the column sums corresponding to each set. Thus PCR4 does a more exact redistribution than both minC versions (versions a) and b) explicated in section 5. %
The PCR4 rule partially extends Dempster's rule in the sense that instead of redistributing the total conflicting mass as within Dempster's rule, PCR4 redistributes partial conflicting masses, hence PCR4 does a better refined redistribution than Dempster's rule; PCR4 and Dempster's rule coincide for $\Theta = \{A, B\}$, in Shafer's model, with $s \geq 2$ sources, and such that $m_{12\ldots s}(A)>0$, $m_{12\ldots s}(B)>0$, and $m_{12\ldots s}(A\cup B)=0$. Thus according to authors opinion, PCR4 rule redistributes better than Dempster's rule since in PCR one goes on partial conflicting, while Dempster's rule redistributes the conflicting mass to all non-empty sets whose conjunctive mass is nonzero, even those not involved in the conflict. 

\subsection{The PCR4 formula}

The PCR4 formula for $s=2$ sources:  $\forall X\in G\setminus\{\emptyset\}$
\begin{equation}
m_{PCR4}(X)=m_{12}(X)\cdot [ 1
+ \sum_{\substack{Y\in G \\ c(Y\cap X)=\emptyset}} \frac{m_{12}(X\cap Y)}{m_{12}(X)+m_{12}(Y)}]
   \label{eq:PCR4}
 \end{equation}
\noindent
with $m_{12}(X)$ and $m_{12}(Y)$ nonzero. $m_{12}(.)$ corresponds to the conjunctive consensus, i.e. $$m_{12}(X)\triangleq \sum_{\substack{X_1,X_2\in G \\ X_1\cap X_2=X}}m_{1}(X_1)m_{2}(X_2)\, .$$
\noindent
If at least one of $m_{12}(X)$ or $m_{12}(Y)$ is zero, the fraction is discarded and the mass $m_{12}(X\cap Y)$ is transferred to $X$ and $Y$ proportionally with respect to their non-zero column sum of masses; if both their column sums of masses are zero, then one transfers to the partial ignorance $X\cup Y$; if even this partial ignorance is empty then one transfers to the total ignorance.\\

Let $G=\{X_1,\ldots,X_n\}\neq \emptyset$ ($G$ being either the power-set or hyper-power set depending on the model we want to deal with), $n\geq 2$, $\forall X\neq\emptyset$, $X\in G$, the general PCR4 formula for $s\geq 2$ sources is given by $\forall X\in G\setminus\{\emptyset\}$
\begin{equation}
m_{PCR4}(X) = m_{12\ldots s}(X)\cdot [1  + \sum_{k=1}^{s-1} S^{PCR4}(X,k)]
\label{eq:PCR4s}
\end{equation}
\noindent
with
\begin{equation}
S^{PCR4}(X,k)\triangleq
\sum_{\substack{X_{i_1},\ldots, X_{i_k}\in G\setminus\{X\}\\
\{i_1,\ldots,i_k\} \in \mathcal{P}^k(\{1,2,\ldots,n\})
 \\ c(X\cap X_{i_1}\cap \ldots \cap X_{i_k})=\emptyset}} 
 \frac{m_{12\ldots s}(X\cap X_{i_1}\cap \ldots \cap X_{i_k})}{m_{12\ldots s}(X)+\sum_{j=1}^{k} m_{12\ldots s}(X_{i_j})}
\end{equation}

\noindent
with all $m_{12\ldots s}(X)$, $m_{12\ldots s}(X_1)$, \ldots, $m_{12\ldots s}(X_n)$ nonzero and where the first term of the right side of \eqref{eq:PCR4s} corresponds to the conjunctive consensus between $s$ sources (i.e. $m_{12\ldots s}(.)$). If at least one of $m_{12\ldots s}(X)$, $m_{12\ldots s}(X_1)$, \ldots, $m_{12\ldots s}(X_n)$ is zero, the fraction is discarded 
and the mass $m_{12\ldots s}(X\cap X_1\cap X_2\cap\ldots\cap X_k)$ is transferred to $X$, $X_1$, \ldots, $X_k$ proportionally with respect to their corresponding column sums in the mass matrix.

\subsection{Example for PCR4 versus minC}
Let's consider $\Theta=\{A,B\}$, Shafer's model  and the the two following bbas:
$$m_1(A)=0.6 \quad m_1(B)=0.3 \quad m_1(A\cup B)=0.1$$
$$m_2(A)=0.2 \quad m_2(B)=0.3 \quad m_2(A\cup B)=0.5$$
Then the conjunctive consensus yields :
$$m_{12}(A)=0.44 \quad m_{12}(B)=0.27 \quad m_{12}(A\cup B)=0.05$$
with the conflicting mass
$$k_{12}=m_{12}(A\cap B)=m_1(A)m_2(B)+m_1(B)m_2(A)=0.24$$

Applying PCR4 rule, one has the following proportional redistribution\footnote{$x$ is the part of conflict redistributed to $A$, $y$ is the part of conflict redistributed to $B$.}  to satisfy
$$\frac{x}{0.44}=\frac{y}{0.27}=\frac{0.24}{0.44+0.27}\approx 0.3380$$
\noindent from which, one deduces $x=0.1487$ and $y=0.0913$ and thus
\begin{align*}
& m_{PCR4}(A) = 0.44 + 0.1487 = 0.5887\\
& m_{PCR4}(B) = 0.27 + 0.0913 = 0.3613\\
& m_{PCR4}(A\cup B) = 0.05
\end{align*}

\noindent
while applying minC (version a) and b) are equivalent in this 2D case), one uses the following proportional redistribution\footnote{$z$ is the part of conflict redistributed to $A\cup B$.} 
$$\frac{x}{0.44}=\frac{y}{0.27}=\frac{z}{0.05}=\frac{0.24}{0.44+0.27+0.05}\approx 0.31578$$
\noindent
Whence $x=0.44\cdot (0.24/0.76) \approx 0.138947$, $y = 0.27 \cdot(0.24/0.76) \approx 0.085263$, $z = 0.05\cdot  (0.24/0.76) \approx 0.015789$, so that
\begin{align*}
& m_{minC}(A) \approx 0.44+ 0.138947 = 0.578948\\
& m_{minC}(B) \approx 0.27+ 0.085263 = 0.355263\\
& m_{minC}(A\cup B) \approx 0.05+ 0.015789 = 0.065789
\end{align*}

Therefore, one sees clearly the difference between PCR4 and minC rules. It can be noted here that 
minC gives the same result as Dempster's rule, but the result drawn from minC and Dempster's rules is less exact in comparison to PCR4 because minC and  Dempster's rules redistribute a fraction of the conflicting mass to $A\cup B$ too, although $A\cup B$ is not involved in any conflict (therefore $A\cup B$ doesn't deserve anything).\\

We can remark also that in the 2D Bayesian case, the PCR4, minC, and Dempster's rules give the same results. For example, let's take $\Theta=\{A,B\}$, the Shafer's model and the two following bbas
$$m_1(A)=0.6 \quad m_1(B)=0.4$$
$$m_2(A)=0.1 \quad m_2(B)=0.9$$
The conjunctive consensus yields $m_{12}(A)=0.06$, $m_{12}(B)=0.36$ with the conflicting mass
$$k_{12}=m_{12}(A\cap B)=m_1(A)m_2(B)+m_1(B)m_2(A)=0.58$$
PCR4, MinC and Dempster's rules provide 
\begin{align*}
& m_{PCR4}(A)=m_{minC}(A)=m_{DS}(A)= 0.142857\\
& m_{PCR4}(B)=m_{minC}(B)=m_{DS}(B)= 0.857143
\end{align*}

\subsection{Example of neutral impact of VBA for PCR4}

Let's consider the previous example with $\Theta=\{A,B\}$, Shafer's model  and the the two following bbas:
$$m_1(A)=0.6 \quad m_1(B)=0.3 \quad m_1(A\cup B)=0.1$$
$$m_2(A)=0.2 \quad m_2(B)=0.3 \quad m_2(A\cup B)=0.5$$
Then the conjunctive consensus yields :
$$m_{12}(A)=0.44 \quad m_{12}(B)=0.27 \quad m_{12}(A\cup B)=0.05$$
with the conflicting mass
$$k_{12}=m_{12}(A\cap B)=m_1(A)m_2(B)+m_1(B)m_2(A)=0.24$$

The canonical form $c( A\cap B ) = A\cap B$, thus $k_{12}=m_{12}( A\cap B ) = 0.24$ will be distributed to $A$ and $B$ only proportionally with respect to their corresponding $m_{12}(.)$, i.e. with respect to 0.44 and 0.27 respectively. One gets:
$$m_{PCR4|12}(A)=0.5887   \quad  m_{PCR4|12}(B)=0.3613 \quad m_{PCR4|12}(A\cup B)=0.05$$
Now let's introduce a third and vacuous belief assignment $m_v(A\cup B)=1$ and combine altogether 
$m_1(.)$, $m_2(.)$ and $m_v(.)$ with the conjunctive consensus. One gets
$$m_{12v}(A)=0.44 \quad m_{12v}(B)=0.27 \quad m_{12v}(A\cup B)=0.05 \quad m_{12v}(A\cap B \cap (A\cup B))=0.24$$
Since the  canonical form $c( A\cap B\cap (A\cup B) ) = A\cap B$, $m_{12v}( A\cap B \cap (A\cup B)) = 0.24$ will be distributed to $A$ and $B$ only proportionally with respect to their corresponding $m12v(.)$, i.e. with respect to 0.44 and 0.27 respectively, therefore exactly as above. Thus
$$m_{PCR4|12v}(A)=0.5887   \quad  m_{PCR4|12v}(B)=0.3613 \quad m_{PCR4|12v}(A\cup B)=0.05$$
\noindent
In this example one sees that the neutrality property of VBA is effectively well satisfied by PCR4 rule since
$$m_{PCR4|12v}(.)=m_{PCR4|12}(.)$$
\noindent
A general proof for neutrality of VBA within PCR4 is given in section \ref{SectionPCR5}.

\subsection{A more complex example for PCR4}

Let's consider now a more complex example involving some null masses (i.e. $m_{12}(A)=m_{12}(B)=0$ ) in the conjunctive consensus between sources. So, let's consider $\Theta=\{A,B,C,D\}$, the Shafer's model and the two following belief assignments:
$$m_1(A)=0\quad m_1(B)=0.4\quad m_1(C)=0.5 \quad m_1(D)=0.1$$
$$m_2(A)=0.6\quad m_2(B)=0\quad m_2(C)=0.1 \quad m_2(D)=0.3$$
The conjunctive consensus yields here $m_{12}(A)=m_{12}(B)=0$, $m_{12}(C)=0.05$, $m_{12}(D)=0.03$ with the total conflicting mass
\begin{align*}
k_{12}&=m_{12}(A\cap B)+m_{12}(A\cap C)+m_{12}(A\cap D)  + m_{12}(B\cap C)+ m_{12}(B\cap D) + m_{12}(C\cap D)\\
&= 0.24+ 0.30 + 0.06 + 0.04 + 0.12 + 0.16=0.92
\end{align*}

Because $m_{12}(A)=m_{12}(B)=0$, the denominator $m_{12}(A)+m_{12}(B)=0$ and the transfer onto $A$ and $B$ should be done proportionally to $m_2(A)$ and $m_1(B)$, thus:
$$\frac{x}{0.6}=\frac{y}{0.4}=\frac{0.24}{0.6+0.4}=0.24$$
\noindent whence $x = 0.144$, $y=0.096$.\\

\noindent 
$m_{12}(A\cap C)=0.30$ is transferred to $A$ and $C$:
$$\frac{x}{0.6}=\frac{z}{0.5}=\frac{0.30}{1.1}$$
\noindent Hence $x=0.6\cdot (0.30/1.1)=0.163636$ and $z=0.5\cdot (0.30/1.1)=0.136364$.\\

\noindent 
$m_{12}(A\cap D)=0.06$ is transferred to $A$ and $D$:
$$\frac{x}{0.6}=\frac{w}{0.4}=\frac{0.06}{1}$$
\noindent Hence $x=0.06\cdot (0.06)=0.036$ and $w=0.4\cdot (0.06)=0.024$.\\

\noindent 
$m_{12}(B\cap C)=0.06$ is transferred to $B$ and $C$:
$$\frac{y}{0.4}=\frac{z}{0.6}=\frac{0.04}{1}$$
\noindent Hence $y=0.4\cdot (0.04)=0.016$ and $z=0.6\cdot (0.04)=0.024$.\\

\noindent 
$m_{12}(B\cap D)=0.06$ is transferred to $B$ and $D$:
$$\frac{y}{0.4}=\frac{w}{0.4}=\frac{0.12}{0.8}=0.15$$
\noindent Hence $y=0.4\cdot (0.15)=0.06$ and $w=0.4\cdot (0.15)=0.06$.\\

The partial conflict $m_{12}(C\cap  D)=0.16$ is proportionally redistributed to $C$ and $D$ only according to
$$\frac{z}{0.05}=\frac{w}{0.03}=\frac{0.16}{0.05+0.03}=2$$
\noindent whence $z=0.10$ and $w=0.06$. Summing all redistributed partial conflicts, one finally gets:
\begin{align*}
m_{PCR4}(A)&=0+0.144+0.163636+0.036= 0.343636\\
m_{PCR4}(B)&= 0+0.096+0.016+0.016=0.172000\\
m_{PCR4}(C)&=0.05+0.136364+0.024+0.10=0.310364\\
m_{PCR4}(D)&=0.03+0.024+0.06+0.06=0.174000
\end{align*}

\noindent
while minC provides\footnote{It can be proven that versions a) and b) of minC provide here same result because in this specific example $m_{12}(A)=m_{12}(B)=m_{12}(A\cup B)=0$.}
$$m_{minC}(A)=m_{minC}(B)=m_{minC}(A\cup B)=0.08 \qquad m_{minC}(C)= 0.490  \qquad m_{minC}(D)=0.270$$
The distinction between PCR4 and minC here is that minC transfers  equally the 1/3 of conflicting mass $m_{12}(A\cap B)=0.24$ onto $A$, $B$ and $A\cup B$, while PCR4 redistributes it to $A$ and $B$ proportionally to their masses $m_2(A)$ and $m_1(B)$. Upon to authors opinions, the minC redistribution appears less exact than PCR4 since $A\cup B$  is not involved into the partial conflict $A\cap B$ and we don't see a reasonable justification on minC transfer onto $A\cup B$ in this case.

\section{The PCR5 rule}
\subsection{Principle of PCR5}
\label{SectionPCR5}
Similarly to PCR2-4, PCR5 redistributes the partial conflicting mass to the elements involved in the partial conflict, considering the canonical form of the partial conflict. PCR5 is the most mathematically exact redistribution of conflicting mass to non-empty sets
following the logic of the conjunctive rule. But this is harder to implement. 
PCR5 satisfies the neutrality property of VBA also. In order to understand the principle of PCR5, letÕs start with examples going from the easiest to the more complex one.\\

\noindent
{\it{Proof of neutrality of VBA for PCR2-PCR5}}: PCR2, PCR3, PCR4 and PCR5 rules preserve the neutral impact of the VBA because in any partial conflict, as well in the total conflict which is a sum of all partial conflicts, the canonical form of each partial conflict does not include $\Theta$ since $\Theta$ is a neutral element for intersection (conflict), therefore $\Theta$ gets no mass after the redistribution of the conflicting mass. This general proof for neutrality of VBA works in dynamic or static cases for all PCR2-5, since the total ignorance, say $I_t$, can not escape the conjunctive normal form, i.e. $c(I_t\cap A)=A$, where $A$ is any set included in $D^\Theta$.

\subsubsection{A two sources example 1 for PCR5}

Suppose one has the frame of discernment $\Theta=\{A, B\}$ of exclusive elements, and 2 sources
of evidences providing the following bbas
$$m_1(A)=0.6 \quad m_1(B)=0 \quad m_1(A\cup B)=0.4$$
$$m_2(A)=0 \quad m_2(B)=0.3 \quad m_2(A\cup B)=0.7$$
Then the conjunctive consensus yields :
$$m_{12}(A)=0.42 \quad m_{12}(B)=0.12 \quad m_{12}(A\cup B)=0.28$$
with the conflicting mass
$$k_{12}=m_{12}(A\cap B)=m_1(A)m_2(B)+m_1(B)m_2(A)=0.18$$
Therefore $A$ and $B$ are involved in the conflict ($A\cup B$ is not involved), hence only $A$ and $B$ deserve a part of the conflicting mass, $A\cup B$ does not deserve. With PCR5, one redistributes the conflicting mass 0.18 to $A$ and $B$ proportionally with the masses $m_1(A)$
and $m_2(B)$ assigned to $A$ and $B$ respectively. Let $x$ be the conflicting mass to be redistributed
to $A$, and $y$ the conflicting mass redistributed to $B$, then
$$\frac{x}{0.6}=\frac{y}{0.3}=\frac{x+y}{0.6+0.3}=\frac{0.18}{0.9}=0.2$$
\noindent
whence $x = 0.6\cdot 0.2 = 0.12$, $y = 0.3\cdot 0.2 = 0.06$. Thus:
\begin{align*}
m_{PCR5}(A) &= 0.42 + 0.12 = 0.54\\
m_{PCR5}(B) &= 0.12 + 0.06 = 0.18\\
m_{PCR5}(A\cup B) &= 0.28
\end{align*}
This result is equal to that of PCR3 and even PCR2, but different from PCR1 and PCR4 in this specific example. PCR1 and PCR4 yield:
\begin{align*}
m_{PCR1}(A) &= 0.42 + \frac{0.6+0}{2}\cdot 0.18 = 0.474\\
m_{PCR1}(B) &= 0.12 + \frac{0+0.3}{2}\cdot 0.18  = 0.147\\
m_{PCR1}(A\cup B) &= 0.28 + \frac{0.4+0.7}{2}\cdot 0.18 = 0.379\\
&\\
m_{PCR4}(A) &= 0.42 + 0.42\cdot\frac{0.18}{0.42+0.12} = 0.56\\
m_{PCR4}(B) &= 0.12 +  0.12\cdot\frac{0.18}{0.12+0.42}  = 0.16\\
m_{PCR4}(A\cup B) &= 0.28
\end{align*}

\subsubsection{A two sources example 2 for PCR5}

Now let's modify a little the previous example and consider now:
$$m_1(A)=0.6 \quad m_1(B)=0 \quad m_1(A\cup B)=0.4$$
$$m_2(A)=0.2 \quad m_2(B)=0.3 \quad m_2(A\cup B)=0.5$$
Then the conjunctive consensus yields :
$$m_{12}(A)=0.50 \quad m_{12}(B)=0.12 \quad m_{12}(A\cup B)=0.20$$
with the conflicting mass
$$k_{12}=m_{12}(A\cap B)=m_1(A)m_2(B)+m_1(B)m_2(A)=0.18$$

The conflict $k_{12}$ is the same as in previous example, which means that $m_2(A) = 0.2$ did not have any impact on the conflict; why?, because $m_1(B) = 0$.
Therefore $A$ and $B$ are involved in the conflict ($A\cup B$ is not involved), hence only $A$ and $B$ deserve a part of the conflicting mass, $A\cup B$ does not deserve.
With PCR5, one redistributes the conflicting mass 0.18 to $A$ and $B$ proportionally with the masses $m_1(A)$ and  $m_2(B)$ assigned to $A$ and $B$ respectively. The mass $m_2(A) = 0.2$ is not considered to the weighting factors of the redistribution. Let $x$ be the conflicting mass to be redistributed to
$A$, and $y$ the conflicting mass redistributed to $B$. By the same calculations one has:
$$\frac{x}{0.6}=\frac{y}{0.3}=\frac{x+y}{0.6+0.3}=\frac{0.18}{0.9}=0.2$$
\noindent
whence $x = 0.6\cdot 0.2 = 0.12$, $y = 0.3\cdot 0.2 = 0.06$. Thus, one gets now:
\begin{align*}
m_{PCR5}(A) &= 0.50 + 0.12 = 0.62\\
m_{PCR5}(B) &= 0.12 + 0.06 = 0.18\\
m_{PCR5}(A\cup B) &= 0.20 +0 = 0.20
\end{align*}

We did not take into consideration the sum of masses of column $A$, i.e.
$m_1(A)+m_2(A) = 0.6 + 0.2 = 0.8$, since clearly $m_2(A) = 0.2$ has no impact on the conflicting
mass.\\

\noindent
In this second example, the result obtained by PCR5 is different from WAO, PCR1, PCR2, PCR3 and PCR4 because
\begin{align*}
m_{WAO}(A) &= 0.50 + \frac{0.6+0.2}{2}\cdot 0.18 = 0.572\\
m_{WAO}(B) &= 0.12 + \frac{0+0.3}{2}\cdot 0.18  = 0.147\\
m_{WAO}(A\cup B) &= 0.20 + \frac{0.4+0.5}{2}\cdot 0.18 = 0.281\\
&\\
m_{PCR1}(A) &= 0.50 + \frac{0.6+0.2}{0.8+0.3+0.9}\cdot 0.18 = 0.572\\
m_{PCR1}(B) &= 0.12 + \frac{0+0.3}{0.8+0.3+0.9}\cdot 0.18  = 0.147\\
m_{PCR1}(A\cup B) &= 0.20 + \frac{0.4+0.5}{0.8+0.3+0.9}\cdot 0.18 = 0.281\\
& \\
m_{PCR2}(A) &= 0.50 + \frac{0.6+0.2}{0.8+0.3}\cdot 0.18 \approx 0.631\\
m_{PCR2}(B) &= 0.12 + \frac{0+0.3}{0.8+0.3}\cdot 0.18  \approx 0.169\\
m_{PCR2}(A\cup B) &= 0.20\\
& \\
m_{PCR3}(A)& = 0.50 + 0.8\cdot [\frac{0.6\cdot 0.3 + 0.2\cdot 0}{0.8 + 0.3}] \approx 0.631\\
m_{PCR3}(B) &= 0.12 + 0.3\cdot [\frac{0.6\cdot 0.3 + 0.2\cdot 0}{0.8 + 0.3}] \approx 0.169\\
m_{PCR3}(A\cup B) &= 0.20\\
& \\
m_{PCR4}(A) &= 0.50 + 0.50\cdot\frac{0.18}{0.50+0.12} \approx 0.645\\
m_{PCR4}(B) &= 0.12 +  0.12\cdot\frac{0.18}{0.50+0.12}  \approx 0.155\\
m_{PCR4}(A\cup B) &= 0.20
\end{align*}

Let's examine from this example the convergence of the PCR5 result by introducing a small positive increment on $m_1(B)$, i.e. one starts now with the PCR5 combination of the following bbas
\begin{align*}
m_1(A)&=0.6 &\quad m_1(B)&=\epsilon &\quad m_1(A\cup B)&=0.4-\epsilon\\
m_2(A)&=0.2 &\quad m_2(B)&=0.3 &\quad m_2(A\cup B)&=0.5
\end{align*}
Then the conjunctive consensus yields: $m_{12}(A)=0.50-0.2\cdot\epsilon$, $m_{12}(B)=0.12+0.5\cdot\epsilon$, $m_{12}(A\cup B)=0.20-0.5\cdot\epsilon$
with the conflicting mass
\begin{align*}
k_{12}&=m_{12}(A\cap B)=m_1(A)m_2(B)+m_1(B)m_2(A)=0.18+0.2\cdot\epsilon
\end{align*}

\noindent
Applying the PCR5 rule for $\epsilon=0.1$, $\epsilon=0.01$,$\epsilon=0.001$ and $\epsilon=0.0001$ one gets the following result:
\begin{table}[h]
\begin{center}
\begin{tabular}{|c||l|l|l|}
\hline
 $\epsilon$     & $m_{PCR5}(A)$ & $m_{PCR5}(B)$ & $m_{PCR5}(A\cup B)$  \\ 
\hline 
\hline
0.1    & 0.613333 & 0.236667 & 0.15\\
0.01  & 0.619905 & 0.185095 & 0.195\\
0.001  & 0.619999 & 0.180501 & 0.1995\\
0.0001  & 0.62 & 0.180050 & 0.19995\\
\hline
\end{tabular}
\end{center}
\caption{Convergence of PCR5}
\label{TableContinuity}
\end{table}

From Table \ref{TableContinuity}, one can see that when $\epsilon$ tend towards zero, the results tends towards the previous result $m_{PCR5}(A)=0.62$, $m_{PCR5}(B)=0.18$ and $m_{PCR5}(A\cup B)=0.20$. Let's explain now in details how this limit can be achieved formally. With PCR5, one redistributes the partial conflicting mass 0.18 to $A$ and $B$ proportionally with the masses $m_1(A)$ and $m_2(B)$ assigned to $A$ and $B$ respectively, and also the partial conflicting mass $0.2\cdot\epsilon$ to $A$ and $B$ proportionally with the masses $m_2(A)$ and $m_1(B)$ assigned to $A$ and $B$ respectively, thus one gets now two weighting factors in the redistribution for each
corresponding set $A$ and $B$. Let $x_1$ be the conflicting mass to be redistributed to $A$, and $y_1$ the conflicting mass redistributed to $B$ from the first partial conflicting mass 0.18. This first partial proportional redistribution is then done according
$$\frac{x_1}{0.6}=\frac{y_1}{0.3}=\frac{x_1+y_1}{0.6+0.3}=\frac{0.18}{0.9}=0.2$$
\noindent
whence $x_1 = 0.6\cdot 0.2 = 0.12$, $y_1 = 0.3\cdot 0.2 = 0.06$. 
Now let $x_2$ be the conflicting mass to be redistributed to $A$, and $y_2$ the conflicting mass
redistributed to $B$ from the second partial conflicting mass $0.2\cdot\epsilon$. This first partial proportional redistribution is then done according
$$\frac{x_2}{0.2}=\frac{y_2}{\epsilon}=\frac{x_2+y_2}{0.2+\epsilon}=\frac{0.2\cdot\epsilon}{0.2+\epsilon}$$
\noindent
whence $x_2 = 0.2\cdot \frac{0.2\cdot\epsilon}{0.2+\epsilon}$, $y_2 =\epsilon\frac{0.2\cdot\epsilon}{0.2+\epsilon}$. 
Thus one gets the following result
\begin{align*}
m_{PCR5}(A) &= m_{12}(A) + x_1+x_2= (0.50-0.2\cdot\epsilon) + 0.12 + 0.2\cdot \frac{0.2\cdot\epsilon}{0.2+\epsilon}\\
m_{PCR5}(B) &= m_{12}(B) + y_1 + y_2=(0.12+0.5\cdot\epsilon) + 0.06 + \epsilon\frac{0.2\cdot\epsilon}{0.2+\epsilon}\\
m_{PCR5}(A\cup B) &= m_{12}(A\cup B)=0.20-0.5\epsilon
\end{align*}
\noindent
From these formal expressions of $m_{PCR5}(.)$, one sees directly that
$$\lim_{\epsilon\rightarrow 0}  m_{PCR5}(A)=0.62 \qquad \lim_{\epsilon\rightarrow 0}  m_{PCR5}(B)=0.18 \qquad \lim_{\epsilon\rightarrow 0}  m_{PCR5}(A\cup B)=0.20$$

\subsubsection{A two sources example 3 for PCR5}
\label{EX3PCR5}

Let's go further modifying this time the previous example and considering:
$$m_1(A)=0.6 \quad m_1(B)=0.3 \quad m_1(A\cup B)=0.1$$
$$m_2(A)=0.2 \quad m_1(B)=0.3 \quad m_1(A\cup B)=0.5$$
Then the conjunctive consensus yields :
$$m_{12}(A)=0.44 \quad m_{12}(B)=0.27 \quad m_{12}(A\cup B)=0.05$$
with the conflicting mass
\begin{align*}
k_{12}&=m_{12}(A\cap B)=m_1(A)m_2(B)+m_1(B)m_2(A)=0.18+0.06=0.24
\end{align*}
The conflict $k_{12}$ is now different from the two previous examples, which means that $m_2(A) = 0.2$ and $m_1(B) =0.3$ did make an impact on the conflict; why?, because $m_2(A)m_1(B) = 0.2\cdot 0.3 = 0.06$ was added to the conflicting mass. Therefore $A$ and $B$ are involved in the conflict ($A\cup B$ is not involved), hence only $A$ and $B$ deserve a
part of the conflicting mass, $A\cup B$ does not deserve.
With PCR5, one redistributes the partial conflicting mass 0.18 to $A$ and $B$ proportionally with the masses $m_1(A)$ and $m_2(B)$ assigned to $A$ and $B$ respectively, and also the partial conflicting mass 0.06 to $A$ and $B$ proportionally with the masses $m_2(A)$ and $m_1(B)$ assigned to $A$ and $B$ respectively, thus one gets two weighting factors of the redistribution for each
corresponding set $A$ and $B$ respectively.
Let $x_1$ be the conflicting mass to be redistributed to $A$, and $y_1$ the conflicting mass redistributed to $B$ from the first partial conflicting mass 0.18. This first partial proportional redistribution is then done according
$$\frac{x_1}{0.6}=\frac{y_1}{0.3}=\frac{x_1+y_1}{0.6+0.3}=\frac{0.18}{0.9}=0.2$$
\noindent
whence $x_1 = 0.6\cdot 0.2 = 0.12$, $y_1 = 0.3\cdot 0.2 = 0.06$. 
Now let $x_2$ be the conflicting mass to be redistributed to $A$, and $y_2$ the conflicting mass
redistributed to $B$ from second the partial conflicting mass 0.06. This second partial proportional redistribution is then done according
$$\frac{x_2}{0.2}=\frac{y_2}{0.3}=\frac{x_2+y_2}{0.2+0.3}=\frac{0.06}{0.5}=0.12$$
\noindent
whence $x_2 = 0.2\cdot 0.12 = 0.024$, $y_2 = 0.3\cdot 0.12 = 0.036$. 
Thus:
\begin{align*}
m_{PCR5}(A) &= 0.44 + 0.12 + 0.024 = 0.584\\
m_{PCR5}(B) &= 0.27 + 0.06 + 0.036 = 0.366\\
m_{PCR5}(A\cup B) &= 0.05 + 0 = 0.05
\end{align*}
\noindent
The result is different from PCR1, PCR2, PCR3 and PCR4 since one has\footnote{The verification is left to the reader.}:
\begin{align*}
m_{PCR1}(A) &= 0.536\\
m_{PCR1}(B) &= 0.342\\
m_{PCR1}(A\cup B) &= 0.122\\
& \\
m_{PCR2}(A) &= m_{PCR3}(A) \approx 0.577\\
m_{PCR2}(B) &= m_{PCR3}(B) \approx 0.373\\
m_{PCR2}(A\cup B) &= m_{PCR3}(A\cup B) =0.05\\
& \\
m_{PCR4}(A) &\approx 0.589\\
m_{PCR4}(B) &\approx 0.361\\
m_{PCR4}(A\cup B) &= 0.05
\end{align*}
The Dempster's rule, denoted here by index DS, gives for this example:
$$m_{DS}(A) = \frac{0.44}{1-0.24} \approx 0.579\qquad
m_{DS}(B) = \frac{0.27}{1-0.24}\approx 0.355\qquad
m_{DS}(A\cup B) = \frac{0.05}{1-0.24}\approx 0.066$$
One clearly sees that $m_{DS}(A\cup B)$ gets some mass from the conflicting mass although $A\cup B$ does not deserve any part of the conflicting mass since $A\cup B$ is not involved in the conflict (only $A$ and $B$ are involved in the conflicting mass). Dempster's rule appears to authors opinions less exact than PCR5 because it redistribute less exactly the conflicting mass than PCR5, even than PCR4 and minC, since Dempter's rule takes the total conflicting mass and redistributes it to all non-empty sets, even those not involved in the conflict.

\subsection{The PCR5 formula}

Before explaining the general procedure to apply for PCR5 (see next section), we give here the PCR5 formula for $s=2$ sources:  $\forall X\in G\setminus\{\emptyset\}$
\begin{equation}
m_{PCR5}(X)=m_{12}(X) 
+\sum_{\substack{Y\in G\setminus\{X\} \\ c(X\cap Y)=\emptyset}} 
[\frac{m_1(X)^2m_2(Y)}{m_1(X)+m_2(Y)} + \frac{m_2(X)^2 m_1(Y)}{m_2(X)+m_1(Y)}]
   \label{eq:PCR5}
 \end{equation}
\noindent
where $c(x)$ represents the canonical form of $x$, $m_{12}(.)$ corresponds to the conjunctive consensus, i.e. $m_{12}(X)\triangleq \sum_{\substack{X_1,X_2\in G \\ X_1\cap X_2=X}}m_{1}(X_1)m_{2}(X_2)$ and where all denominators are {\it{different from zero}}. If a denominator is zero, that fraction is discarded.\\

\noindent
Let $G=\{X_1,\ldots,X_n\}\neq \emptyset$ ($G$ being either the power-set or hyper-power set depending on the model we want to deal with), $n\geq 2$, the general PCR5 formula for $s\geq 2$ sources is given by $\forall X\in G\setminus\{\emptyset\}$
\begin{multline}
m_{PCR5}(X) = m_{12\ldots s}(X)+ \sum_{\substack{2\leq t\leq s\\ 1\leq r_1,\ldots,r_t\leq s\\
1\leq r_1< r_2 < \ldots < r_{t-1} < (r_t=s)}}
 \sum_{\substack{X_{j_2},\ldots,X_{j_t}\in G\setminus\{X\}\\
\{j_2,\ldots,j_t\} \in \mathcal{P}^{t-1}(\{1,\ldots,n\})\\
c(X\cap X_{j_2}\cap \ldots\cap X_{j_s})=\emptyset\\
\{i_1,\ldots,i_s\} \in \mathcal{P}^{s}(\{1,\ldots,s\})}}\\
\frac{(\prod_{k_1=1}^{r_1} m_{i_{k_1}}(X)^2)\cdot [\prod_{l=2}^{t}( \prod_{k_l=r_{l-1} +1}^{r_l} m_{i_{k_l}}(X_{j_l})]}{(\prod_{k_1=1}^{r_1} m_{i_{k_1}}(X)) + [\sum_{l=2}^{t}( \prod_{k_l=r_{l-1} +1}^{r_l} m_{i_{k_l}}(X_{j_l})]}
\label{eq:PCR5s}
\end{multline}

\noindent
where  $i$, $j$, $k$, $r$, $s$ and $t$ in \eqref{eq:PCR5s} are integers. $m_{12\ldots s}(X)$ corresponds to the conjunctive consensus on $X$ between $s$ sources and where all denominators are different from zero. If a denominator is zero, that fraction is discarded; $\mathcal{P}^{k}(\{1,2,\ldots,n\})$ is the set of all subsets of $k$ elements from $\{1,2,\ldots,n\}$ (permutations of $n$ elements taken by $k$), the order of elements doesn't count.\\

Let's prove here that \eqref{eq:PCR5s} reduces to \eqref{eq:PCR5} when $s=2$. Indeed, if one takes $s=2$ in general PCR5 formula \eqref{eq:PCR5s},let's note first that: 
\begin{itemize}
\item
$2\leq t\leq s$ becomes $2\leq t\leq 2$, thus $t=2$.
\item $1 \leq r_1, r_2 \leq (s=2)$, or $r_1, r_2 \in \{1,2\}$, 
but because $r_1 < r_2$ one gets $r_1=1$ and $r_2=2$.
\item $m_{12\ldots s}(X)$ becomes $m_{12}(X)$
\item $X_{j_2},\ldots,X_{j_t}\in G\setminus\{X\}$ becomes $X_{j_2}\in G\setminus\{X\}$ because $t=2$.
\item $\{j_2,\ldots,j_t\} \in \mathcal{P}^{t-1}(\{1,\ldots,n\})$ becomes $j_2\in \mathcal{P}^{1}(\{1,\ldots,n\})=\{1,\ldots,n\}$
\item $c(X\cap X_{j_2}\cap \ldots\cap X_{j_s})=\emptyset$ becomes $c(X\cap X_{j_2})=\emptyset$
\item $\{i_1,\ldots,i_s\} \in \mathcal{P}^{s}(\{1,\ldots,s\})$ becomes $\{i_1,i_2\} \in \mathcal{P}^{2}(\{1,2\})=\{\{1,2\},\{2,1\}\}$
\end{itemize}

\noindent
Thus  \eqref{eq:PCR5s}  becomes when $s=2$,
\begin{equation*}
m_{PCR5}(X) = m_{12}(X) + \sum_{\substack{t=2\\  r_1=1, r_2=2}}
 \sum_{\substack{X_{j_2}\in G\setminus\{X\}\\
j_2 \in \{1,\ldots,n\}\\
c(X\cap X_{j_2})=\emptyset\\
\{i_1,i_2\} \in\{\{1,2\},\{2,1\}\}}}
\frac{(\prod_{k_1=1}^{1} m_{i_{k_1}}(X)^2)\cdot [\prod_{l=2}^{2}( \prod_{k_l=r_{l-1}+1}^{r_l} m_{i_{k_l}}(X_{j_l})]}{(\prod_{k_1=1}^{1} m_{i_{k_1}}(X)) + [\sum_{l=2}^{2}( \prod_{k_l=r_{l-1} +1}^{r_l} m_{i_{k_l}}(X_{j_l})]}
\label{eq:PCR5s1}
\end{equation*}

\noindent
After elementary algebraic simplification, it comes
\begin{equation*}
m_{PCR5}(X) = m_{12}(X) +
 \sum_{\substack{X_{j_2}\in G\setminus\{X\}\\
j_2 \in \{1,\ldots,n\}\\
c(X\cap X_{j_2})=\emptyset\\
\{i_1,i_2\} \in\{\{1,2\},\{2,1\}\}}}
\frac{m_{i_{1}}(X)^2\cdot [ \prod_{k_2=2}^{2} m_{i_{k_2}}(X_{j_2}]}{ m_{i_{1}}(X) + [ \prod_{k_2=2}^{2} m_{i_{k_2}}(X_{j_2}]}
\label{eq:PCR5s2}
\end{equation*}

\noindent
Since $\prod_{k_2=2}^{2} m_{i_{k_2}}(X_{j_2})=m_{i_{2}}(X_{j_2})$ and condition "$X_{j_2}\in G\setminus\{X\}$ and $j_2 \in \{1,\ldots,n\}$" are equivalent to $X_{j_2}\in G\setminus\{X\}$, one gets:
\begin{equation*}
m_{PCR5}(X) = m_{12}(X) +
 \sum_{\substack{X_{j_2}\in G\setminus\{X\}\\
c(X\cap X_{j_2})=\emptyset\\
\{i_1,i_2\} \in\{\{1,2\},\{2,1\}\}}}
\frac{m_{i_{1}}(X)^2\cdot  m_{i_{2}}(X_{j_2})}{ m_{i_{1}}(X) + m_{i_{2}}(X_{j_2})}
\label{eq:PCR5s3}
\end{equation*}

\noindent
This formula can also be written as (denoting $X_{j_2}$ as $Y$)
\begin{equation*}
m_{PCR5}(X) = m_{12}(X) +
 \sum_{\substack{Y \in G\setminus\{X\}\\
c(X\cap Y)=\emptyset}}
[\frac{m_{1}(X)^2  m_{2}(Y)}{ m_{1}(X) + m_{2}(Y)} +
\frac{m_{2}(X)^2 m_{1}(Y)}{ m_{2}(X) + m_{1}(Y)}
]
\label{eq:PCR5s4}
\end{equation*}

\noindent
which is the same as formula \eqref{eq:PCR5}. 
Thus the proof is completed.

\subsection{General procedure to apply the PCR5}

Here is the general procedure to apply PCR5:
\begin{enumerate}
\item apply the conjunctive rule;
\item calculate all partial conflicting masses separately;
\item if $A\cap B = \emptyset$ then $A$, $B$ are involved in the conflict; redistribute the mass
$m_{12}(A\cap B)>0$ to the non-empty sets $A$ and $B$ proportionally with respect to 
\begin{enumerate}
\item[a)] the non-zero masses $m_1(A)$ and $m_2(B)$ respectively,
\item[b)] the non-zero masses $m_2(A)$ and $m_1(B)$ respectively, and
\item[c)] other non-zero masses that occur in some products of the
sum of $m_{12}(A\cap B)$;
\end{enumerate}
\item
if both sets $A$ and $B$ are empty, then the transfer is forwarded to the disjunctive form
$u(A)\cup u(B)$, and if this disjunctive form is also empty, then the transfer is
forwarded to the total ignorance in a closed world (or to the empty set if the open
world approach is preferred); but if even the total ignorance is empty one considers an open world (i.e. new
hypotheses might exist) and the transfer is forwarded to the empty set;
if say $m_1(A) = 0$ or $m_2(B) = 0$, then the product $m_1(A)m_2(B) = 0$ and thus there is no
conflicting mass to be transferred from this product to non-empty sets;
if both products $m_1(A)m_2(B) = m_2(A)m_1(B) = 0$ then there is no conflicting mass to
be transferred from them to non-empty sets;
in a general case\footnote{An easier calculation method, denoted {\it{PCR5-approximate}} for $s \geq 3$ bbas, which is an {\it{approximation of PCR5}}, is to first combine $s-1$ bbas altogether using the conjunctive rule, and the result to be again combined once more with the $s$-th bba also using the conjunctive rule; then the weighting factors will only depend on $m_{12\ldots (s-1)}(.)$ and $m_s(.)$ only - instead of depending on all bbas $m_1(.)$, $m_2(.)$, \ldots, $m_s(.)$. PCR5-approximate result however depends on the chosen order of the sources.}
, for $s \geq 2$ sources, the mass $m_{12\ldots s}(A_1\cap A_2\cap\ldots\capÉA_r)>0$, with $2 \leq r \leq s$, where $A_1\cap A_2\cap \ldots \cap A_r = \emptyset$, resulted from the application of the conjunctive rule,
is a sum of many products; each non-zero particular product is proportionally
redistributed to $A_1$, $A_2$, \ldots, $A_r$ with respect to the sub-products of masses assigned to
$A_1$, $A_2$, \ldots, $A_r$ respectively by the sources;
 if both sets $A_1$, $A_2$, \ldots, $A_r$ are empty, then the transfer is forwarded to the disjunctive
form $u(A_1)\cup u(A_2)\cup\ldots\cup u(A_r)$, and if this disjunctive form is also empty, then the
transfer is forwarded to the total ignorance in a closed world (or to the empty set if the open
world approach is preferred); but if even the total ignorance is empty one considers an open world
(i.e. new hypotheses might exist) and the transfer is forwarded to the empty set;
\item and so on until all partial conflicting masses are redistributed;
\item add the redistributed conflicting masses to each corresponding non-empty set
involved in the conflict;
\item the sets not involved in the conflict do not receive anything from the conflicting
masses (except some partial or total ignorances in degenerate cases).
\end{enumerate}

The more hypotheses and more masses are involved in the fusion, the more difficult is to
implement PCR5. Yet, it is easier to {\it{approximate PCR5}} by first combining $s-1$ bbas through the conjunctive rule, then by combining again the result with the $s$-th bba also using the conjunctive rule Ð in order to reduce very much the calculations of the redistribution of conflicting mass.

\subsection{A 3 sources example for PCR5}

Let's see a more complex example using PCR5. Suppose one has the frame of discernment $\Theta=\{A, B\}$ of exclusive elements, and 3 sources such that:
$$m_1(A)=0.6 \quad m_1(B)=0.3 \quad m_1(A\cup B)=0.1$$
$$m_2(A)=0.2 \quad m_2(B)=0.3 \quad m_2(A\cup B)=0.5$$
$$m_3(A)=0.4 \quad m_3(B)=0.4 \quad m_3(A\cup B)=0.2$$
Then the conjunctive consensus yields :
$m_{123}(A)=0.284$, $m_{123}(B)=0.182$ and $m_{123}(A\cup B)=0.010$
with the conflicting mass $k_{123}=m_{123}(A\cap B)=0.524$, which is a sum of factors.
\begin{enumerate}
\item {\it{Fusion based on PCR5}}:\\

In the long way, each product occurring as a term in the sum of the conflicting mass
should be redistributed to the non-empty sets involved in the conflict proportionally to the
masses (or sub-product of masses) corresponding to the respective non-empty set.
For example, the product $m_1(A)m_3(B)m_2(A\cup B) = 0.6\cdot 0.4\cdot 0.5 = 0.120$ occurs in the sum of $k_{123}$, then $0.120$ is proportionally distributed to the sets involved in the conflict; because $c(A\cap B\cap (A\cup B))=A\cap B$ the transfer is done to $A$ and $B$ with respect to to $0.6$ and $0.4$. Hence:
$$\frac{x}{0.6} = \frac{y}{0.4}= \frac{0.12}{0.6+0.4} $$
\noindent
whence $x = 0.6\cdot 0.12 = 0.072$, $y = 0.4\cdot 0.12 = 0.048$, which will be added to the masses of $A$ and $B$ respectively.
Another example, the product $m_2(A)m_1(B)m_3(B) = 0.2\cdot 0.3\cdot 0.4 = 0.024$ occurs in the sum
of $k_{123}$, then $0.024$ is proportionally distributed to $A$, $B$ with respect to $0.20$ and
$0.3\cdot 0.4=0.12$ respectively. Hence:
$$\frac{x}{0.20} = \frac{y}{0.12} = \frac{0.024}{0.32} = 0.075$$
whence $x = 0.20\cdot\frac{0.024}{0.32} = 0.015$ and $y = 0.12\cdot \frac{0.024}{0.32} = 0.009$, which will be added to the masses of $A$, and $B$ respectively.\\

\noindent
But this procedure is more difficult, that's why we can use the following crude approach:
\item {\it{Fusion based on PCR5-approximate}}:\\

\noindent
If $s$ sources are involved in the fusion, then first combine using the conjunctive rule $s-1$
sources, and the result will be combined with the remaining source.
 \end{enumerate}
 
\noindent
We resolve now this 3 sources example by combining the first two sources
$$m_1(A)=0.6 \quad m_1(B)=0.3 \quad m_1(A\cup B)=0.1$$
$$m_2(A)=0.2 \quad m_2(B)=0.3 \quad m_2(A\cup B)=0.5$$
with the DSm classic rule (i.e. the conjunctive consensus on hyper-power set $D^\Theta$) to get
$$m_{12}(A)=0.44 \quad m_{12}(B)=0.27$$
$$m_{12}(A\cup B)=0.05\quad m_{12}(A\cap B)=0.24$$
Then one combines $m_{12}(.)$ with $m_3(.)$ still with the DSm classic rule and one gets as preliminary step for PCR5-version b just above-mentioned
$$m_{123}(A)=0.284 \qquad m_{123}(B)=0.182$$
$$m_{123}(A\cup B)=0.010\quad m_{123}(A\cap B)=0.524$$
The conflicting mass has been derived from 
\begin{align*}
m_{123}(A\cap B) & = [ m_{12}(A)m_3(B) + m_3(A)m_{12}(B) ]  + [ m_3(A)m_{12}(A\cap B) + m_3(B)m_{12}(A\cap B) \\
& \quad + m_3(A\cap B)m_{12}(A\cap B) ] \\
&= [ 0.44\cdot 0.4 + 0.4\cdot 0.27 ]  + [ 0.4\cdot 0.24 + 0.4\cdot 0.24 + 0.2\cdot 0.24 ]=0.524
\end{align*}
But in the last brackets $A\cap B=\emptyset$, therefore the masses of $m_3(A)m_{12}(A\cap B) = 0.096$, $m_3(B)m_{12}(A\cap B) = 0.096$, and $m_3(A\cap B)m_{12}(A\cap B) = 0.048$ are transferred to $A$, $B$, and $A\cup B$ respectively.
In the first brackets, $0.44\cdot 0.4 = 0.176$ is transferred to $A$ and $B$ proportionally to $0.44$ and
$0.4$ respectively:
$$\frac{x}{0.44} = \frac{y}{0.40} = \frac{0.176}{0.84}$$
\noindent
whence $$x = 0.44 \cdot \frac{0.176}{0.84} = 0.09219\qquad y = 0.40\cdot \frac{0.176}{0.84} = 0.08381$$
\noindent
Similarly, $0.4\cdot 0.27 = 0.108$ is transferred to $A$ and $B$ proportionally to $0.40$ and $0.27$ and one gets:
$$\frac{x}{0.40} = \frac{y}{0.27} = \frac{0.108}{0.67}$$
\noindent
whence 
$$x = 0.40\cdot \frac{0.108}{0.67} = 0.064478\qquad y = 0.27\cdot \frac{0.108}{0.67} = 0.043522$$
Adding all corresponding masses, one gets the final result with PCR5 (version b), denoted here with index $PCR5b|\{12\}3$ to emphasize that one has applied the version b) of PCR5 for the combination of the 3 sources by combining first the sources 1 and 2 together :
$$m_{PCR5b|\{12\}3}(A)=0.536668 \qquad m_{PCR5b|\{12\}3}(B)=0.405332 \qquad m_{PCR5b|\{12\}3}(A\cup B)=0.058000$$

\subsection{On the neutral impact of VBA for PCR5}

Let's take again the example given in section \ref{EX3PCR5} with $\Theta=\{A,B\}$, the Shafer's model and the two bbas
$$m_1(A)=0.6 \quad m_1(B)=0.3 \quad m_1(A\cup B)=0.1$$
$$m_2(A)=0.2 \quad m_1(B)=0.3 \quad m_1(A\cup B)=0.5$$
Then the conjunctive consensus yields :
$$m_{12}(A)=0.44 \quad m_{12}(B)=0.27 \quad m_{12}(A\cup B)=0.05$$
with the conflicting mass
$$k_{12}=m_{12}(A\cap B)=m_1(A)m_2(B)+m_1(B)m_2(A)=0.18+0.06=0.24$$
The canonical form $c(A\cap B) = A\cap B$, thus $m_{12}(A\cap B)= 0.18 + 0.06 = 0.24$ will be distributed to $A$ and $B$ only proportionally with respect to their corresponding masses assigned by $m1(.)$ and $m2(.)$, i.e:
0.18 redistributed to $A$ and $B$ proportionally with respect to 0.6 and 0.3 respectively,
and 0.06 redistributed to $A$ and $B$ proportionally with respect to 0.2 and 0.3 respectively.
One gets as computed above (see also section \ref{EX3PCR5}):
$$m_{PCR5|12}(A) = 0.584 \qquad m_{PCR5|12}(B) =  0.366\qquad m_{PCR5|12}(A\cup B) = 0.05$$

\noindent
Now let's introduce a third and vacuous belief assignment $m_v(A\cup B)=1$ and combine altogether 
$m_1(.)$, $m_2(.)$ and $m_v(.)$ with the conjunctive consensus. One gets
$$m_{12v}(A)=0.44 \quad m_{12v}(B)=0.27 \qquad m_{12v}(A\cup B)=0.05 \qquad m_{12v}(A\cap B \cap (A\cup B))=0.24$$

\noindent
Since the  canonical form $c( A\cap B\cap (A\cup B) ) = A\cap B$, $m_{12v}( A\cap B \cap (A\cup B)) = 0.18 + 0.06 = 0.24$ will be distributed to $A$ and $B$ only (therefore nothing to $A\cup B$) proportionally with respect to their corresponding masses assigned by $m_1(.)$ and $m_2(.)$ (because $m_v(.)$ is not involved since all its masses assigned to $A$ and $B$ are zero: $m_v(A) = m_v(B) = 0$), i.e: 0.18 redistributed to $A$ and $B$ proportionally with respect to 0.6 and 0.3 respectively,
and 0.06 redistributed to $A$ and $B$ proportionally with respect to 0.2 and 0.3 respectively, therefore exactly as above. Thus
$$m_{PCR5|12v}(A)=0.584   \qquad  m_{PCR5|12v}(B)=0.366\qquad m_{PCR5|12v}(A\cup B)=0.05$$
\noindent
In this example one sees that the neutrality property of VBA is effectively well satisfied by PCR5 rule since
$$m_{PCR5|12v}(.)=m_{PCR5|12}(.)$$
\noindent A general proof for neutrality of VBA within PCR5 is given in section \ref{SectionPCR5}.

\section{Numerical examples and comparisons}

In this section, we present some numerical examples and comparisons of PCR rules with other rules proposed in literature.

\subsection{Example 1}

Let's consider the frame of discernment $\Theta=\{A, B, C\}$, Shafer's model (i.e. all intersections
empty), and the 2 following Bayesian bbas
$$m_1(A)=0.6 \quad m_1(B)=0.3 \quad m_1(C)=0.1$$
$$m_2(A)=0.4 \quad m_2(B)=0.4 \quad m_2(C)=0.2$$

Then the conjunctive consensus yields :
$m_{12}(A)=0.24$, $m_{12}(B)=0.12$ and $m_{12}(C)=0.02$
with the conflicting mass $k_{12}=m_{12}(A\cap B) + m_{12}(A\cap C) + m_{12}(B\cap C) = 0.36 + 0.16 + 0.10 = 0.62$, which is a sum of factors.\\

\noindent
From the PCR1 and PCR2 rules, one gets
$$m_{PCR1}(A)=0.550 \qquad m_{PCR2}(A)=0.550$$
$$m_{PCR1}(B)=0.337 \qquad m_{PCR2}(B)=0.337$$
$$m_{PCR1}(C)=0.113 \qquad m_{PCR2}(C)=0.113$$

\noindent
And from the PCR3 and PCR5 rules, one gets
$$m_{PCR3}(A)=0.574842 \qquad m_{PCR5}(A)=0.574571$$
$$m_{PCR3}(B)=0.338235 \qquad m_{PCR5}(B)=0.335429 $$
$$m_{PCR3}(C)=0.086923 \qquad m_{PCR5}(C)=0.090000$$

Dempster's rule is a particular case of proportionalization, where the conflicting mass is
redistributed to the non-empty sets $A_1$, $A_2$, \ldots proportionally to $m_{12}(A_1)$, $m_{12}(A_2)$, \ldots respectively (for the case of 2 sources) and similarly for $n$ sources, i.e.
$$\frac{x}{0.24} = \frac{y}{0.12} = \frac{z}{0.02} = \frac{0.62}{0.38}$$
\noindent
whence $x = 0.24\cdot \frac{0.62}{0.38} = 0.391579$, $y = 0.12\cdot \frac{0.62}{0.38} = 0.195789$, $z =
0.02\cdot \frac{0.62}{0.38} = 0.032632$. The Dempster's rule yields
$$m_{DS}(A) = 0.24 + 0.391579 = 0.631579$$
$$m_{DS}(B) = 0.12 + 0.195789 = 0.315789$$
$$m_{DS}(C) =0.02 + 0.032632 = 0.052632$$

\noindent
Applying PCR4 for this example, one has
$$\frac{x_1}{0.24}=\frac{y_1}{0.12}=\frac{0.36}{0.24+0.12}$$
\noindent therefore $x_1 = 0.24$ and $y_1 = 0.12$;
$$\frac{x_2}{0.24}=\frac{z_1}{0.02}=\frac{0.16}{0.24+0.02}=\frac{0.16}{0.26}$$
\noindent therefore $x_2 = 0.24(0.16/0.26) = 0.147692$ and
$z_1 = 0.02(0.16/0.26) = 0.012308$:
$$\frac{y_2}{0.12}=\frac{z_2}{0.02}=\frac{0.10}{0.12+0.02}=\frac{0.10}{0.14}$$
\noindent therefore $y_2 = 0.12(0.10/0.14) = 0.085714$ and $z_2 = 0.02(0.10/0.14) = 0.014286$.
Summing all of them, one gets finally:
$$m_{PCR4}(A)=0.627692\qquad m_{PCR4}(B)=0.325714\qquad m_{PCR4}(C)=0.046594$$
It can be showed that minC combination provides same result as PCR4 for this example.

\subsection{Example 2}

Let's consider the frame of discernment $\Theta=\{A, B\}$, the Shafer's model (i.e. all intersections empty),
and the following two bbas:
\begin{align*}
m_1(A)&=0.7 &\quad m_1(B)&=0.1 &\quad m_1(A\cup B)&=0.2\\
m_2(A)&=0.5 &\quad m_2(B)&=0.4 &\quad m_2(A\cup B)&=0.1
\end{align*}
Then the conjunctive consensus yields $m_{12}(A)=0.52$, $m_{12}(B)=0.13$ and $m_{12}(A\cup B)=0.02$ with the total conflict $k_{12}=m_{12}(A\cap B)  = 0.33$.\\

\noindent
From PCR1 and PCR2 rules, one gets:
\begin{align*}
m_{PCR1}(A)&=0.7180 & m_{PCR2}(A)&=0.752941\\
m_{PCR1}(B)&=0.2125 & m_{PCR2}(B)&=0.227059\\
m_{PCR1}(A\cup B)&=0.0695 & m_{PCR2}(A\cup B)&=0.02
\end{align*}
\noindent 
From PCR3 and PCR5 rules, one gets
\begin{align*}
m_{PCR3}(A)&=0.752941 & m_{PCR5}(A)&=0.739849\\
m_{PCR3}(B)&=0.227059 & m_{PCR5}(B)&=0.240151\\
m_{PCR3}(A\cup B)&=0.02 & m_{PCR5}(A\cup B)&= 0.02
\end{align*}
\noindent
From the Dempster's rule:
$$m_{DS}(A) = 0.776119 \qquad m_{DS}(B) = 0.194030\qquad m_{DS}(A\cup B) = 0.029851$$
\noindent
From PCR4, one has
$$\frac{x}{0.52}=\frac{y}{0.13}=\frac{0.33}{0.52+0.13}=\frac{0.33}{0.65}$$
\noindent therefore $x = 0.52(0.33/0.65) = 0.264$
and $y = 0.13(0.33/0.65) = 0.066$. Summing, one gets:
$$m_{PCR4}(A)=0.784\qquad m_{PCR4}(B)=0.196\qquad m_{PCR4}(A\cup B)=0.02$$

\noindent
From minC, one has
$$\frac{x}{0.52}=\frac{y}{0.13}=\frac{z}{0.02}=\frac{0.33}{0.52+0.13+0.02}=\frac{0.33}{0.67}$$
\noindent therefore $x = 0.52(0.33/0.67) = 0.256119$, $y = 0.13(0.33/0.67) = 0.064030$ and $z = 0.02(0.33/0.02) = 0.009851$. Summing, one gets same result as with the Demspter's rule in this second example:
$$m_{minC}(A)=0.776119/qquad m_{minC}(B)=0.194030/qquad m_{minC}(A\cup B)=0.029851$$
                
\subsection{Example 3 (Zadeh's example)}

Let's consider the famous Zadeh's example\footnote{A detailed discussion on this example can be found in \cite{DSmTBook_2004a} (Chap. 5, p. 110).} \cite{Zadeh_1979} with $\Theta=\{A,B,C\}$, Shafer's model and the two following belief assignments
\begin{align*}
m_1(A)&=0.9 &\quad m_1(B)&=0 &\quad m_1(C)&=0.1\\
m_2(A)&=0 &\quad m_2(B)&=0.9 &\quad m_2(C)&=0.1
\end{align*}
The conjunctive consensus yields for this case, $m_{12}(A)=m_{12}(b)=0$, $m_{12}(C)=0.01$.
The masses committed to partial conflicts are given by $m_{12}(A\cap B)=0.81$, $m_{12}(A\cap C)=m_{12}(B\cap C)=0.09$
and the conflicting mass by
$$k_{12}=m_1(A)m_2(B)+m_1(A)m_2(C)+m_2(B)m_1(C)=0.81+0.09+0.09=0.99$$

\noindent
The first partial conflict $m_{12}(A\cap B)=0.9\cdot 0.9=0.81$ is proportionally redistributed to $A$ and $B$ according to
$$\frac{x_1}{0.9}=\frac{y_1}{0.9}=\frac{0.81}{0.9+0.9}$$
\noindent
whence $x_1=0.405$ and $y_1=0.405$.\\

The second partial conflict $m_{12}(A\cap C)=0.9\cdot 0.1=0.09$ is proportionally redistributed to $A$ and $C$ according to
$$\frac{x_2}{0.9}=\frac{y_2}{0.1}=\frac{0.09}{0.9+0.1}$$
\noindent
whence $x_2=0.081$ and $y_2=0.009$.\\

The third partial conflict $m_{12}(B\cap C)=0.9\cdot 0.1=0.09$ is proportionally redistributed to $B$ and $C$ according to
$$\frac{x_3}{0.9}=\frac{y_3}{0.1}=\frac{0.09}{0.9+0.1}$$
\noindent
whence $x_3=0.081$ and $y_3=0.009$.\\

\noindent
After summing all proportional redistributions of partial conflicts to corresponding elements with PCR5, one finally gets:
\begin{align*}
m_{PCR5}(A)&=0+0.405 + 0.081 = 0.486\\
m_{PCR5}(B)&=0+0.405 + 0.081 = 0.486\\
m_{PCR5}(C)&=0.01+0.009+0.009 = 0.028
\end{align*}
\noindent
The fusion obtained from other rules yields:

\begin{itemize}
\item with Dempster's rule based on Shafer's model, one gets the counter-intuitive result
$$m_{DS}(C)=1$$
\item with Smets' rule based on Open-World model, one gets 
\begin{equation*}
m_{S}(\emptyset)=0.99\qquad m_{S}(C)=0.01
\end{equation*}
\item with Yager's rule based on Shafer's model, one gets 
\begin{equation*}
m_{Y}(A\cup B \cup C)=0.99\qquad m_{DS}(C)=0.01
\end{equation*}
\item with Dubois \& Prade's rule based on Shafer's model, one gets 
\begin{equation*}
m_{DP}(A\cup B)=0.81\qquad
m_{DP}(A\cup C)=0.09\qquad
m_{DP}(B\cup C)=0.09\qquad
m_{DP}(C)=0.01
\end{equation*}
\item with the classic DSm rule based on the free-DSm model, one gets 
\begin{equation*}
m_{DSmC}(A\cap B)=0.81\qquad
m_{DSmC}(A\cap C)=0.09\qquad
m_{DSmC}(B\cap C)=0.09\qquad
m_{DSmC}(C)=0.01
\end{equation*}
\item with the hybrid DSm rule based on Shafer's model, one gets  same as with Dubois \& Prade (in this specific example)
\begin{equation*}
m_{DSmH}(A\cup B)=0.81\qquad
m_{DSmH}(A\cup C)=0.09\qquad
m_{DSmH}(B\cup C)=0.09\qquad
m_{DSmH}(C)=0.01
\end{equation*}
\item with the WAO rule based on Shafer's model, one gets  
\begin{align*}
m_{WAO}(A)&=0 + \frac{0.9+0}{2}\cdot 0.99=0.4455\\
m_{WAO}(B)&=0 + \frac{0+0.9}{2}\cdot 0.99=0.4455\\
m_{WAO}(C)&=0.01 + \frac{0.1+0.1}{2}\cdot 0.99=0.1090
\end{align*}
\item with the PCR1 rule based on Shafer's model, one gets (same as with WAO)
\begin{align*}
m_{PCR1}(A)&=0 + \frac{0.9}{0.9+0.9+0.2}\cdot 0.99=0.4455\\
m_{PCR1}(B)&=0 + \frac{0.9}{0.9+0.9+0.2}\cdot 0.99=0.4455\\
m_{PCR1}(C)&=0.01 + \frac{0.2}{0.9+0.9+0.2}\cdot 0.99=0.1090
\end{align*}
\item with the PCR2 rule based on Shafer's model, one gets in this example the same result as with WAO and PCR1.
\item with the PCR3 rule based on Shafer's model, one gets
\begin{align*}
m_{PCR3}(A)&=0 + 0.9\cdot [\frac{0\cdot 0 + 0.9\cdot 0.9}{0.9+0.9}+ \frac{0.1\cdot 0 + 0.9\cdot 0.1}{0.9+0.2}]\approx 0.478636\\
m_{PCR3}(B)&=0 + 0.9\cdot [\frac{0\cdot 0 + 0.9\cdot 0.9}{0.9+0.9}+ \frac{0.1\cdot 0 + 0.9\cdot 0.1}{0.9+0.2}]\approx 0.478636\\
m_{PCR3}(C)& \approx 0.042728
\end{align*}
\item
With the PCR4 rule based on Shafer's model,
$m_{12}(A\cap B)=0.81$ is distributed to $A$ and $B$ with respect to their $m_{12}(.)$ masses, but because $m_{12}(A)$ and $m_{12}(B)$ are zero, it is distributed to $A$ and $B$ with respect to their corresponding column sum of masses, i.e. with respect to $0.9+0=0.9$ and $0+0.9=0.9$;
$$\frac{x_1}{0.9}=\frac{y_1}{0.9}= \frac{0.81}{0.09+0.09}$$
\noindent
hence $x_1=0.405$ and $y_1=0.405$.\\

\noindent 
$m(A\cap C)=0.09$ is redistributed to $A$ and $C$ proportionally with respect to their corresponding column sums, i.e. 0.9 and 0.2 respectively:
$$x/0.9=z/0.2=0.09/1.1$$
Hence $x=0.9\cdot (0.09/1.1)=0.073636$ and $z=0.2\cdot (0.09/1.1)=0.016364$.\\

\noindent 
$m(B\cap C)=0.09$ is redistributed to $B$ and $C$ proportionally with respect to their corresponding column sums, i.e. 0.9 and 0.2 respectively:
$$y/0.9=z/0.2=0.09/1.1$$
Hence $y=0.9\cdot (0.09/1.1)=0.073636$ and $z=0.2\cdot (0.09/1.1)=0.016364$.\\

\noindent
Summing one gets:
$$m_{PCR4}(A)=0.478636\qquad m_{PCR4}(B)=0.478636\qquad m_{PCR4}(C)=0.042728$$
\item
With the minC rule based on Shafer's model, one gets:
$$m_{minC}(A)=0.405\qquad m_{minC}(B)=0.405\qquad m_{minC}(C)=0.190$$

\item With the PCR5 rule based on Shafer's model, the mass $m_{12}(A\cap B)=0.9\cdot 0.9=0.81$ is proportionalized according to
$$\frac{x}{0.9}=\frac{y}{0.9}=\frac{0.81}{0.9+0.9}$$
\noindent
whence $x=0.405$ and $y=0.405$.
Similarly, $m_{12}(A\cap C)=0.09$ is proportionalized according to
$$\frac{x}{0.9}=\frac{z}{0.9}=\frac{0.09}{0.9+0.1}$$
\noindent whence $x = 0.081$ and $z = 0.009$;
Similarly, $m_{12}(B\cap C) = 0.09$ is proportionalized according to
$$\frac{y}{0.9}=\frac{z}{0.1}=\frac{0.09}{0.9+0.1}$$
\noindent whence $y = 0.081$ and $z = 0.009$.
Summing one gets:
\begin{align*}
m_{PCR5}(A)&=0+0.405 + 0.081 = 0.486\\
m_{PCR5}(B)&=0+0.405 + 0.081 = 0.486\\
m_{PCR5}(C)&=0.01+0.009+0.009 = 0.028
\end{align*}
\end{itemize}

\subsection{Example 4 (hybrid model)}
Let's consider a hybrid model on $\Theta=\{A,B,C\}$ where $A\cap B=\emptyset$, while 
$A\cap C\neq \emptyset$ and $B\cap C\neq \emptyset$. This model corresponds to a hybrid model \cite{DSmTBook_2004a}.  Then only the mass $m_{12}(A\cap B)$ of partial conflict $A\cap B$ will be transferred to other non-empty sets, while the masses $m_{12}(A\cap C)$ stays on $A\cap C$ and 
$m_{12}(B\cap C)$ stays on $B\cap C$. Let's consider two sources of evidence with the following basic belief assignments
$$m_1(A)=0.5\quad m_1(B)=0.4 \quad m_1(C)=0.1$$
$$m_2(A)=0.6\quad m_2(B)=0.2 \quad m_2(C)=0.2$$
Using the table representation, one has\\

\begin{center}
\begin{tabular}[h]{|c|cccccc|}
\hline
 & $A$ &$ B$ & $C$ & $A\cap B$ & $A\cap C$ & $B\cap C$ \\
 \hline
 $m_1$ & 0.5 & 0.4 & 0.1 & & & \\
 $m_2$ & 0.6 & 0.2 & 0.2 & & & \\
\hline
 $m_{12}$ & 0.3 & 0.08 & 0.02 & 0.34 & 0.16 & 0.10\\
\hline
\end{tabular}\\
\end{center}

\noindent
Thus, the conjunctive consensus yields
$$m_{12}(A)=0.30\qquad m_{12}(B)=0.08 \qquad m_{12}(C)=0.02$$
$$m_{12}(A\cap B)=0.34\qquad m_{12}(A\cap C)=0.16\qquad m_{12}(B\cap C)=0.10$$

\begin{itemize}
\item with the PCR1 rule, $m_{12}(A\cap B)=0.34$  is the only conflicting mass, and it is redistributed to 
$A$, $B$ and $C$ proportionally with respect to their corresponding columns' sums: $0.5+0.6=1.1$, $0.4+0.2=0.6$ and $0.1+0.2=0.3$.
The sets $A\cap C$ and $B\cap C$ don't get anything from the conflicting mass 0.34 since 
their columns' sums are zero. According to proportional conflict redistribution of PCR1, one has
$$\frac{x}{1.1}=\frac{y}{0.6}=\frac{z}{0.3}=\frac{0.34}{1.1+0.6+0.3}=0.17$$
\noindent
Therefore, one gets the proportional redistributions for $A$, $B$ and $C$
$$x=1.1\cdot 0.17=0.187\qquad y=0.6\cdot 0.17=0.102\qquad z=0.3\cdot 0.17=0.051$$
\noindent
Thus the final result of PCR1 is given by
\begin{align*}
m_{PCR1}(A)&=0.30+0.187=0.487\\
m_{PCR1}(B)&=0.08+0.102=0.182\\
m_{PCR1}(C)&=0.02+0.051=0.071\\
m_{PCR1}(A\cap C)&=0.16\\
m_{PCR1}(B\cap C)&=0.10
\end{align*}

\item with the PCR2 rule, $m_{12}(A\cap B)=0.34$  is redistributed to 
$A$ and $B$ only with respect to their corresponding columns' sums: $0.5+0.6=1.1$ and $0.4+0.2=0.6$.
The set $C$ doesn't get anything since $C$ was not involved in the conflict. According to proportional conflict redistribution of PCR2, one has
$$\frac{x}{1.1}=\frac{y}{0.6}=\frac{0.34}{1.1+0.6}=0.2$$
\noindent
Therefore, one gets the proportional redistributions for $A$ and $B$
$$x=1.1\cdot 0.2=0.22\qquad y=0.6\cdot 0.2=0.12$$
\noindent
Thus the final result of PCR2 is given by
\begin{align*}
m_{PCR2}(A)&=0.30+0.22=0.52\\
m_{PCR2}(B)&=0.08+0.12=0.20\\
m_{PCR2}(C)&=0.02\\
m_{PCR2}(A\cap C)&=0.16\\
m_{PCR2}(B\cap C)&=0.10
\end{align*}
\item PCR3 gives the same result like PCR2 since there is only a partial 
conflicting mass which coincides with the total conflicting mass.
\item
with the PCR4 rule, $m_{12}(A\cap B)=0.34$  is redistributed to 
$A$ and $B$ proportionally with respect to $m_{12}(A)=0.30$ and $m_{12}(B)=0.08$. According to proportional conflict redistribution of PCR4, one has
$$\frac{x}{0.30}=\frac{y}{0.08}=\frac{0.34}{0.30+0.08}$$
\noindent
Therefore, one gets the proportional redistributions for $A$ and $B$
$$x=0.30\cdot (0.34/0.38) \approx 0.26842\qquad y=0.08\cdot (0.34/0.38)\approx 0.07158$$
\noindent
Thus the final result of PCR4 is given by
\begin{align*}
m_{PCR4}(A)&=0.30+0.26842=0.56842\\
m_{PCR4}(B)&=0.08+0.07158=0.15158\\
m_{PCR4}(C)&=0.02\\
m_{PCR4}(A\cap C)&=0.16\\
m_{PCR4}(B\cap C)&=0.10
\end{align*}
\item
with the PCR5 rule, $m_{12}(A\cap B)=0.34$  is redistributed to 
$A$ and $B$ proportionally with respect to 
$m_1(A)=0.5$, $m_2(B)=0.2$ and then with respect to $m_2(A)=0.6$, $m_1(B)=0.4$. According to proportional conflict redistribution of PCR5, one has
$$\frac{x_1}{0.5}=\frac{y_1}{0.2}=\frac{0.10}{0.5+0.2}=0.10/0.7\qquad \qquad \frac{x_2}{0.6}=\frac{y_2}{0.4}=\frac{0.24}{0.6+0.4}=0.24$$
\noindent
Therefore, one gets the proportional redistributions for $A$ and $B$
\begin{align*}
x_1&=0.5\cdot (0.10/0.7)=0.07143 &\quad y_1&=0.2\cdot (0.10/0.7)=0.02857\\
x_2&=0.6\cdot 0.24=0.144 &\quad y_2&=0.4\cdot 0.24=0.096
\end{align*}

\noindent
Thus the final result of PCR5 is given by
\begin{align*}
m_{PCR5}(A)&=0.30+0.07143+0.144=0.51543\\
m_{PCR5}(B)&=0.08+0.02857+0.096=0.20457\\
m_{PCR5}(C)&=0.02\\
m_{PCR5}(A\cap C)&=0.16\\
m_{PCR5}(B\cap C)&=0.10
\end{align*}
\end{itemize}

\subsection{Example 5 (Target ID tracking)}

This example is drawn from Target ID (identification) tracking application pointed out by Dezert and al. in \cite{Dezert_2005}. The problem consists in updating bba on ID of a target based on a sequence of uncertain attribute measurements expressed as sensor's bba. In such case, a problem can arise when the fusion rule of the predicted ID bba with the current observed ID bba yields to commit certainty on a given ID of the frame $\Theta$ (the set of possible target IDs under consideration). If this occurs once, then the ID bba remains inchanged by all future observations, whatever the value they can take ! By example, at a given time the ID system finds with "certainty" that a target is a truck, and then during next, say 1000 scans, all the sensor reports claim with high belief that target is a car, but the ID system is unable to doubt itself of his previous ID assessment (certainty state plays actually the role of an absorbing/black hole state). Such behavior of a fusion rule is what we feel drastically dangerous, specially in defence applications and better rules than the classical ones have to be used to avoid such severe drawback. We provide here a simple numerical example and we compare the results for the new rules presented in this paper. So lets consider here the Shafer's model, a 2D frame $\Theta=\{A,B\}$ and two bba $m_1(.)$ and $m_2(.)$ with
\begin{center}
\begin{tabular}[h]{|c|ccc|}
\hline
 & $A$ &$ B$ & $A\cup B$ \\
 \hline
 $m_1$ & 1 & 0 &  0  \\
 $m_2$ & 0.1 & 0.9 &  0 \\
\hline
\end{tabular}\\
\end{center}
\noindent
$m_1(.)$ plays here the role of a prior (or predicted) target ID bba for a given time step and $m_2(.)$ is the observed target ID bba drawn from some attribute measurement for the time step under consideration. The conjunctive operator of the prior bba and the observed bba is then
$$m_{12}(A)=0.1\qquad m_{12}(A\cap B)=0.9$$
Because we are working with Shafer's model, one has to redistribute the conflicting mass $m_{12}(A\cap B)=0.9$ in some manner onto the non conflicting elements of power-set. Once the fusion/update is obtained at a given time, we don't keep in memory $m_1(.)$ and $m_2(.)$ but we only use the fusion result as new prior\footnote{For simplicity, we don't introduce a prediction ID model here and we just consider as predicted bba for time $k+1$, the updated ID bba available at time $k$ (i.e. the ID state transition matrix equals identity matrix).} bba for the fusion with the next observation, and this process is reitered at every observation time. Let's examine the result of the rule after at first observation time (when only $m_2(.)$ comes in).

\begin{itemize}
\item {\bf{With minC rule}}: 
minC rule distributes the whole conflict to $A$ since $m_{12}(B)=0$, thus:
$$m_{minC|12}(A)=1$$
\item {\bf{With PCR1-PCR4 rules}}: 
Using PCR1-4, they all coincide here. One has $x/1.1=y/0.9=0.9/2=0.45$, whence $x=1.1\cdot (0.45)=0.495$ and $y=0.9\cdot (0.45)=0.405$. Hence
$$m_{PCR1-4|12}(A)=0.595\qquad m_{PCR1-4|12}(B)=0.405$$
\item {\bf{With PCR5 rule}}: 
One gets $x/1=y/0.9 = 0.9/1.9$, whence $x=1\cdot (0.9/1.9)=0.473684$ and $y=0.9\cdot (0.9/1.9)=0.426316$. Hence
$$m_{PCR5|12}(A)=0.573684 \qquad m_{PCR5|12}(B)=0.426316$$
\end{itemize}

Suppose a new observation, expressed by $m_3(.)$ comes in at next scan with
$$m_3(A)=0.4\qquad m_3(B)=0.6$$
\noindent
and examine the result of the new target ID bba update based on the fusion of the previous result with $m_3(.)$.

\begin{itemize}
\item {\bf{With minC rule}}: 
The conjunctive operator applied on $m_{minC|12}(.)$ and $m_3(.)$ yields now
$$m_{(minC|12)3}(A)=0.4\qquad m_{(minC|12)3}(A\cap B)=0.6$$
\noindent
Applying minC rule again, one distributes the whole conflict 0.6 to $A$ and one finally gets\footnote{For convenience, we use the notation $m_{minC|(12)3}(A)$ instead of $m_{minC|(minC|12)3}(.)$, and similarly with PCR indexes.}:
$$m_{minC|(12)3}(A)=1$$
\noindent
Therefore, minC rule does not respond to the new tracking ID observations.
\item {\bf{With PCR1-PCR4 rules}}: 
The conjunctive operator applied on $m_{PCR1-4|12}(.)$ and $m_3(.)$ yields now
$$m_{(PCR1-4|12)3}(A)=0.238\quad m_{(PCR1-4|12)3}(B)=0.243\quad m_{(PCR1-4|12)3}(A\cap B)=0.519$$
\begin{itemize}
\item {\bf{For PCR1-3}}: $x/0.995=y/1.005 = 0.519/2=0.2595$, so that
$x=0.995\cdot (0.2595)= 0.258203$ and $y=1.005\cdot (0.2595)=0.260797$.
Hence:
$$m_{PCR1-3|(12)3}(A)= 0.496203 \quad m_{PCR1-3|(12)3}(B)= 0.503797$$
Therefore PCR1-3 rules do respond to the new tracking ID observations.
\item {\bf{For PCR4}}: $x/0.238=y/0.243=0.519/(0.238+0.243)=0.519/0.481$, so that
$x=0.238\cdot (0.519/0.481)=0.256802$ and $y=0.243\cdot (0.519/0.481)=0.262198$.
Hence:
$$m_{PCR4|(12)3}(A)= 0.494802 \quad m_{PCR4|(12)3}(B)= 0.505198$$
\end{itemize}
Therefore PCR4 rule does respond to the new tracking ID observations.
\item {\bf{With PCR5 rule}}: 
The conjunctive operator applied on $m_{PCR5|12}(.)$ and $m_3(.)$ yields now
$$m_{(PCR5|12)3}(A)=0.229474\quad
m_{(PCR5|12)3}(B)= 0.255790\quad
m_{(PCR5|12)3}(A\cap B)=0.514736$$
Then: $x/0.573684=y/0.6=(0.573684\cdot 0.6)/( 0.573684 + 0.6)=0.293273$, so that
$x=0.573684 \cdot 0.293273 = 0.168246$ and $y=0.6 \cdot 0.293273 = 0.175964$.
Also:
$x/0.4=y/0.426316=(0.4 \cdot 0.426316)/( 0.4 + 0.426316)=0.206369$, so that
$x=0.4 \cdot  0.206369 = 0.082548$ and $y=0.426316 á 0.206369 = 0.087978$.
Whence:
$$m_{PCR5|(12)3}(A)=0.480268\quad m_{PCR5|(12)3}(B)= 0.519732$$
Therefore PCR5 rule does respond to the new tracking ID observations.

\end{itemize}

It can moreover be easily verified that the Dempster's rule gives the same results as minC here, hence does not respond to new observations in target ID tracking problem.

\section{On Ad-Hoc-ity of fusion rules}

Each fusion rule is more or less ad-hoc. Same thing for PCR rules. There is up to the
present no rule that fully satisfies everybody. LetÕs analyze some of them.\\

{\it{Dempster's rule}} transfers the conflicting mass to non-empty sets proportionally with their
resulting masses. What is the reasoning for doing this? Just to swell the masses of non-empty
sets in order to sum up to 1?\\

{\it{Smets' rule}} transfers the conflicting mass to the empty set. Why? Because, he says, we
consider on open world where unknown hypotheses might be. This approach does not make difference between all origins of conflicts since all different conflicting masses are committed with the same manner to the empty set. Not convincing. And what about real closed worlds?\\

{\it{Yager's rule}} transfers all the conflicting mass only to the total ignorance. Should the internal structure of partial conflicting
mass be ignored?\\

{\it{Dubois-Prade's rule}} and {\it{DSm hybrid rule}} transfer the conflicting mass to the partial an
total ignorances upon the principle that between two conflicting hypotheses one is right.
Not completely justified either. What about the case when no hypothesis is right?\\

{\it{PCR rules}} are based on total or partial conflicting masses, transferred to the corresponding
sets proportionally with respect to some functions (weighting coefficients) depending on
their corresponding mass matrix columns. But other weighting coefficients can be found.\\

Inagaki \cite{Inagaki_1991}, Lef\`evre-Colot-Vannoorenberghe \cite{Lefevre_2002} proved that there are infinitely many fusion rules based on the conjunctive rule and then on the transfer of the conflicting
mass, all of them depending on the weighting coefficients/factors that transfer that
conflicting mass. How to choose them, what parameters should they rely on Ð thatÕs the
question! There is not a precise measure for this. In authors' opinion, neither DSm hybrid rule nor PCR rules are not more ad-hoc than other fusion rules.

\section{Conclusion}

We have presented in this article five versions of the Proportional Conflict Redistribution rule of
combination in information fusion, which are implemented as follows: first one uses the
conjunctive rule, then one redistribute the conflicting mass to non-empty sets proportionally
with respect to either the non-zero column sum of masses (for PCR1, PCR2, PCR3) or
with respect to the non-zero masses (of the corresponding non-empty set) that enter in the
composition of each individual product in the partial conflicting masses (PCR5). PCR1 restricted from the hyper-power set to the power set and without degenerate cases gives the same result as WAO as pointed out by P. Smets in a private communication. PCR1 and PCR2 redistribute the total conflicting mass, while PCR3 and PCR5 redistribute partial conflicting masses.  PCR1-3 uses the proportionalization with respect to the sum of mass columns, PCR4 with respect to the results of the conjunctive rule, and PCR5 with respect to the masses entered in the sum products of 
the conflicting mass. PCR4 is an improvement of minC and Dempster's rules. From PCR1 to PCR2, PCR3, PCR4, PCR5 one increases the complexity of the rules and also the exactitude of the redistribution of conflicting masses. All the PCR rules proposed in this paper preserve the neutral impact of the vacuous belief assignment but PCR1 and work for any hybrid DSm model (including the Shafer's model). For the free DSm model, i.e. when all intersections not empty, there is obviously no need for transferring any mass since there is no conflicting mass, the masses of the intersections stay on them.  Thus only DSm classic rule is applied, no PCR1-5, no DSm hybrid rule and no other rule needed to apply. In this paper, PCR, minC and Dempster's rules are all compared with respect to the conjunctive rule (i.e. the conjunctive rule is applied first, then the conflicting mass is redistributed following the way the conjunctive rule works). Therefore, considering the way each rule works, the rule which works closer to the conjunctive rule in redistributing the conflicting mass is considered better than other rule.  This is not a subjective comparison between rules, but only a mathematical one.

\section*{Acknowledgements}

We want to thank Dr. Wu Li from NASA Langley Research Center, USA and Prof. Philippe Smets
from the Universit\'e Libre de Bruxelles, Belgium for their interesting discussions during preliminary stage of this work. We are also very grateful to Dr. Milan Daniel from Institute of Computer Science, Prague, Czech Republic, for his fruitful comments and criticisms and his deep examination of this paper.


\begin{thebibliography}{99}

\bibitem{Daniel98x}
Daniel M., \emph{Distribution of Contradictive Belief Masses in Combination of Belief Functions},
Information, Uncertainty and Fusion, Eds. Bouchon-Meunier B., Yager R.~R., and Zadeh L.~A.,
Kluwer Academic Publishers,
2000, pp. 431-446.

\bibitem{Daniel_2000}
Daniel M., \emph{Associativity in Combination of Belief Functions}, 
Proceedings of 5th Workshop on Uncertainty Processing (WUPES 2000), Ed. Vejnarov{\'a}, J., 
Edi{\v c}n{\'\i} odd{\v e}len{\'\i} V{\v S}E (Editorial Department of University of Economics), 2000, pp. 41--54.

\bibitem{Daniel_2000b}
Daniel M., \emph{Associativity and Contradiction in Combination of Belief Functions},
Proceedings  Eighth International conference IPMU, Universidad Politecnica de Madrid, 2000, Vol. I., pp. 133--140.

\bibitem{Daniel_2003}
Daniel M., \emph{Associativity in Combination of belief functions; a derivation of minC combination}, 
Soft Computing, 7(5), pp. 288--296, 2003.

\bibitem{Dezert_2005}
Dezert J., Tchamova A., Semerdjiev T., Konstantinova P., \emph{Performance Evaluation of Fusion Rules for Multitarget Tracking in Clutter based on Generalized Data Association}, 
Submitted to Fusion 2005 Int. Conf., Philadelphia, July 2005 (available upon request to authors).

\bibitem{Dubois_1986b}
Dubois D., Prade H., \emph{A Set-Theoretic View of Belief Functions}, International Journal of General Systems, pp.193-226, Vol.12, 1986.

\bibitem{Dubois_1988}
Dubois D., Prade H., \emph{Representation and combination of uncertainty with belief functions and possibility measures}, Computational Intelligence, 4, pp. 244-264, 1988.

\bibitem{Dubois_1992}
Dubois D., Prade H., \emph{On the combination of evidence in various mathematical frameworks}, Reliability Data Collection and Analysis, J. Flamm and T. Luisi, Brussels, ECSC, EEC, EAFC: pp. 213-241, 1992.

\bibitem{Inagaki_1991}
Inagaki T., \emph{Interdependence between safety-control policy and multiple-sensor schemes via Dempster-Shafer theory}, IEEE Trans. on reliability, Vol. 40, no. 2, pp. 182-188, 1991.

\bibitem{Josang_2003}
J{\o}sang, A., Daniel, M., Vannoorenberghe, P., \emph{Strategies for Combining Conflicting Dogmatic Beliefs}, Proceedings of the 6th International Conference on International Fusion, Vol. 2, pp. 1133-1140, 2003.

\bibitem{Lefevre_2002}
Lef\`evre E., Colot O., Vannoorenberghe P., \emph{Belief functions combination and conflict management}, Information Fusion Journal, Elsevier Publisher, Vol. 3, No. 2, pp. 149-162, 2002.

\bibitem{Murphy_2000}
Murphy C.K., \emph{Combining belief functions when evidence conflicts}, 
Decision Support Systems, Elsevier Publisher, Vol. 29, pp. 1-9, 2000.

\bibitem{Sentz_2002}
Sentz K., Ferson S., \emph{Combination of evidence in Dempster-Shafer Theory}, SANDIA Tech. Report, SAND2002-0835, 96 pages, April 2002. (http://www.sandia.gov/epistemic/Reports/SAND2002-0835.pdf)

\bibitem{Shafer_1976}
Shafer G., \emph{A Mathematical Theory of Evidence}, Princeton Univ. Press, Princeton, NJ, 1976.

\bibitem{DSmTBook_2004a}
Smarandache F., Dezert J. (Editors), \emph{Applications and Advances of DSmT for Information Fusion}, Am. Res. Press, Rehoboth, 2004, 
http://www.gallup.unm.edu/{\verb+~+}smarandache/DSmT-book1.pdf. 

\bibitem{Smarandache_Dezert_2004c}
Smarandache F., Dezert J.,  \emph{A Simple Proportional Conflict Redistribution Rule}, arXiv
Archives, Los Alamos National Laboratory, July-August 2004; the Abstract and the whole paper are available at http://arxiv.org/abs/cs.AI/0408010 and at http://arxiv.org/PS{\verb+_+}cache/cs/pdf/0408/0408010.pdf.

\bibitem{Smets}
Smets Ph., \emph{Quantified Epistemic Possibility Theory seen as an hyper Cautious transferable Belief Model}, available at http://iridia.ulb.ac.be/{\verb+~+}psmets

\bibitem{Smets_1990}
Smets Ph., \emph{The combination of evidence in the Transferable Belief Model}, 
IEEE Trans. on Pattern Analysis and Machine Intelligence, Vol. 12, No. 5, pp. 447-458, 1990.

\bibitem{Smets_1993a}
Smets Ph., \emph{Belief functions: the disjunctive rule of combination and the generalized Bayesian theorem}, International Journal of Approximate reasoning, Vol. 9, pp. 1-35, 1993.

\bibitem{Smets_1994}
Smets Ph., Kennes R., \emph{The transferable belief model}, 
Artif. Intel., 66(2), pp. 191-234, 1994.

\bibitem{Smets_2000}
Smets Ph., \emph{Data Fusion in the Transferable Belief Model}, 
Proc. of Fusion 2000 Intern. Conf. on Information Fusion, Paris, July 2000.

\bibitem{Voorbraak_1991}
Voorbraak F., \emph{On the justification of Dempster's rule of combination}, Artificial Intelligence, 48, pp. 171-197, 1991.

\bibitem{Yager_1983}
Yager R. R., \emph{Hedging in the combination of evidence}, Journal of Information and Optimization Science, Vol. 4, No. 1, pp. 73-81, 1983.

 \bibitem{Yager_1985}
Yager R. R., \emph{On the relationships of methods of aggregation of evidence in expert systems}, Cybernetics and Systems, Vol. 16, pp. 1-21, 1985.

\bibitem{Yager_1987}
Yager R.R., \emph{On the Dempster-Shafer framework and new combination rules}, Information Sciences, Vol. 41, pp. 93--138, 1987.

\bibitem{Zadeh_1979}
Zadeh L., \emph{On the validity of Dempster's rule of combination}, Memo M 79/24, Univ. of California, Berkeley, 1979.

\bibitem{Zadeh_1984}
Zadeh L., \emph{Review of Mathematical theory of evidence, by Glenn Shafer}, AI Magazine, Vol. 5, No. 3, pp. 81-83, 1984.

\bibitem{Zadeh_1985}
Zadeh L., \emph{A simple view of the Dempster-Shafer theory of evidence and its implications for the rule of combination}, Berkeley Cognitive Science Report No. 33, University of California, Berkeley, CA, 1985.

\bibitem{Zadeh_1986}
Zadeh L., \emph{A simple view of the Dempster-Shafer theory of evidence and its implication for the rule of combination}, AI Magazine 7, No.2, pp. 85-90, 1986.

\end{thebibliography}
\end{document}